\documentclass[11pt, letterpaper, onecolumn]{tau-class/tau}

\usepackage{amsmath}
\usepackage[english]{babel}
\usepackage{biblatex}
\usepackage{graphicx} 
\usepackage{algorithm}
\usepackage{algpseudocode}
\usepackage{ifthen}
\usepackage{xspace}
\usepackage{siunitx}
\usepackage{wrapfig}
\usepackage{environ}
\usepackage{booktabs}
\usepackage{longtable}
\usepackage{threeparttable}
\usepackage{array}
\usepackage{makecell}
\usepackage{multirow}
\usepackage{multicol}

\newcolumntype{L}{>{\raggedright\arraybackslash\small}p{1.5cm}}
\newcolumntype{J}{>{\raggedright\arraybackslash\small}p{1.5cm}}

\newcommand{\pearl}{\textsc{Pearl}\xspace}

\newcommand{\alphafold}[1]{AlphaFold~#1\xspace}
\newcommand{\af}[1]{AF#1\xspace}
\newcommand{\boltz}[1]{\mbox{Boltz-#1\xspace}}
\newcommand{\chai}[1]{\mbox{Chai-#1\xspace}}
\newcommand{\protenix}{ProteniX\xspace}

\newcommand{\best}[1]{\texttt{best@#1}\xspace}
\newcommand{\bestk}{\best{k}}

\newcommand{\rmsdonea}{\text{RMSD\,\raisebox{0.25ex}{\scriptsize <}\,1\,\AA}\xspace}
\newcommand{\rmsdtwoa}{\text{RMSD\,\raisebox{0.25ex}{\scriptsize <}\,2\,\AA}\xspace}
\newcommand{\rmsdone}{\text{RMSD\,\raisebox{0.25ex}{\scriptsize <}\,1}\xspace}
\newcommand{\rmsdtwo}{\text{RMSD\,\raisebox{0.25ex}{\scriptsize <}\,2}\xspace}
\newcommand{\ltonea}{\text{\raisebox{0.25ex}{\scriptsize <}\,1\,\AA}\xspace}
\newcommand{\lttwoa}{\text{\raisebox{0.25ex}{\scriptsize <}\,2\,\AA}\xspace}

\newcommand{\runsnposes}{Runs~N'~Poses\xspace}

\newcommand{\rnp}{RnP\xspace}

\newcommand{\posebusters}{PoseBusters\xspace}

\newcommand{\internalxtals}{InternalXtals\xspace}

\newcommand{\eg}{\textit{e.g.}\xspace}
\newcommand{\ie}{\textit{i.e.}\xspace}
\let\oldAA\AA
\renewcommand{\AA}{\oldAA\xspace}

\newcommand{\fmtdate}[1]{\mbox{#1\xspace}}
\newcommand{\pval}[1]{\ensuremath{\mathrm{p} \leq #1}}
\newcommand{\cuequivariance}{\texttt{cuEquivariance}\xspace}
\newcommand{\layernorm}{\texttt{LayerNorm}\xspace}

\newcommand{\genesisname}{Genesis Molecular AI\xspace}
\newcommand{\genesislogo}{genesis-logo-new.png}
\newcommand{\logoheight}{1.5cm}

\titleformat{\section}
    {\color{black}\sffamily\large\bfseries}
    {\thesection.}
    {0.5em}
    {#1}
    []
  
\titleformat{name=\section,numberless}
    {\color{black}\sffamily\large\bfseries}
    {}
    {0em}
    {#1}
    []  

\makeatletter
\apptocmd{\ps@firststyle}{\fancyhead[R]{\includegraphics[height=\logoheight]{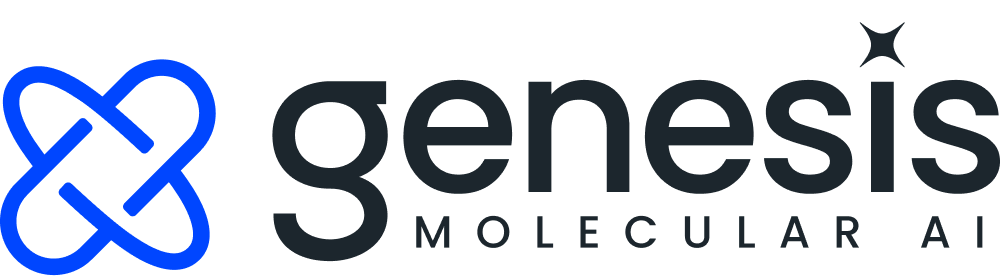}}}{}{}
\makeatother

\journalname{\genesisname\ -- Technical Report}
\title{Pearl: A Foundation Model for Placing Every Atom in the Right Location}

\author[]{Genesis Research Team}
\author[1,$\dagger$]{Alejandro~Dobles}
\author[1,$\dagger$]{Nina~Jovic}
\author[1,$\dagger$]{Kenneth~Leidal}
\author[1,$\dagger$]{Pranav~Murugan}
\author[1,$\dagger$]{David~C.~Williams}
\author[1,$\dagger$]{Drausin~Wulsin}

\author[1,$\ddagger$]{Nate~Gruver}
\author[1,$\ddagger$]{Christina~X.~Ji}
\author[1,$\ddagger$]{Korrawat~Pruegsanusak}
\author[1,$\ddagger$]{Gianluca~Scarpellini}
\author[1,$\ddagger$]{Ansh~Sharma}
\author[1,$\ddagger$]{Wojciech~Swiderski}

\author[1,$\dagger\dagger$]{Andrea~N.~Bootsma}
\author[1,$\dagger\dagger$]{Richard~Strong~Bowen}
\author[1,$\dagger\dagger$]{Charlotte~Chen}
\author[1,$\dagger\dagger$]{Jamin~Chen}
\author[1,$\dagger\dagger$]{Marc~André~Dämgen}
\author[1,$\dagger\dagger$]{Benjamin~DiFrancesco}
\author[1,$\dagger\dagger$]{J.~D.~Fishman}
\author[1,$\dagger\dagger$]{Alla~Ivanova}
\author[1,$\dagger\dagger$]{Zach~Kagin}
\author[1,$\dagger\dagger$]{David~Li-Bland}
\author[1,$\dagger\dagger$]{Zuli~Liu}
\author[1,$\dagger\dagger$]{Igor~Morozov}
\author[1,$\dagger\dagger$,$\mathsection$]{Jeffrey~Ouyang-Zhang}
\author[1,$\dagger\dagger$]{Frank~C.~Pickard~IV}
\author[2,$\dagger\dagger$]{Kushal~S.~Shah}
\author[1,$\dagger\dagger$,$\mathsection$]{Ben~Shor}
\author[1,$\dagger\dagger$]{Gabriel~Monteiro~da~Silva}
\author[2,$\dagger\dagger$]{Roy~Tal}
\author[1,$\dagger\dagger$]{Maxx~Tessmer}
\author[1,$\dagger\dagger$]{Carl~Tilbury}
\author[1,$\dagger\dagger$]{Cyr~Vetcher}
\author[1,$\dagger\dagger$]{Daniel~Zeng}

\author[1,$\#$]{Maruan~Al-Shedivat}
\author[1,$\#$]{Aleksandra~Faust}
\author[1,$\#$]{Evan~N.~Feinberg}
\author[1,$\#$]{Michael~V.~LeVine}
\author[1,$\#$]{Matteus~Pan}

\affil[1]{\genesisname}
\affil[2]{NVIDIA}
\affil[$\dagger$]{Lead Contributor}
\affil[$\ddagger$]{Core Contributor}
\affil[$\dagger\dagger$]{Contributor}
\affil[$\#$]{Senior Contributor}
\affil[$\mathsection$]{Work done during internship}
\professor{Ordered alphabetically within each group}

\institution{\genesisname}
\footinfo{Pearl}
\theday{\today}
\leadauthor{Genesis Molecular AI}

\renewcommand{\absfont}{\selectfont\small\color{black}\sffamily\bfseries}

\begin{abstract}
Accurately predicting the three-dimensional structures of protein–ligand complexes remains a fundamental challenge in computational drug discovery that limits the pace and success of therapeutic design.
Deep learning methods have recently shown strong potential as structural prediction tools, achieving promising accuracy across diverse biomolecular systems.
However, their performance and utility are constrained by scarce experimental data, inefficient architectures, physically invalid poses, and the limited ability to exploit auxiliary information available at inference.
To address these issues, we introduce \pearl (\underline{P}lacing \underline{E}very \underline{A}tom in the \underline{R}ight \underline{L}ocation), a foundation model for protein–ligand cofolding at scale.
\pearl addresses these challenges with three key innovations: (1) training recipes that include large-scale synthetic data to overcome data scarcity; (2) architectures that incorporate an SO(3)-equivariant diffusion module to inherently respect 3D rotational symmetries, improving generalization and sample efficiency, and (3) controllable inference, including a generalized multi-chain templating system supporting both protein and non-polymeric components as well as dual unconditional/conditional modes.
\pearl establishes a new state-of-the-art performance in protein-ligand cofolding.
On the key metric of generating accurate (\rmsdtwoa) and physically valid poses, \pearl surpasses \alphafold3 and other open source baselines on the public \runsnposes and \posebusters benchmarks, delivering 14.5\% and 14.2\% improvements, respectively, over the next best model.
In the pocket-conditional cofolding regime, \pearl delivers $3.6\times$ improvement on a proprietary set of challenging, real-world drug targets at the more rigorous \rmsdonea threshold.
Finally, we demonstrate that model performance correlates directly with synthetic dataset size used in training.
\end{abstract}

\begin{document}

\begin{hyphenrules}{nohyphenation}
\maketitle
\end{hyphenrules}
\thispagestyle{firststyle}

\tauabstract


\section{Introduction}
\label{sec:introduction}

\begin{figure}[!ht]
    \centering
    \includegraphics[keepaspectratio,width=1.0\linewidth]{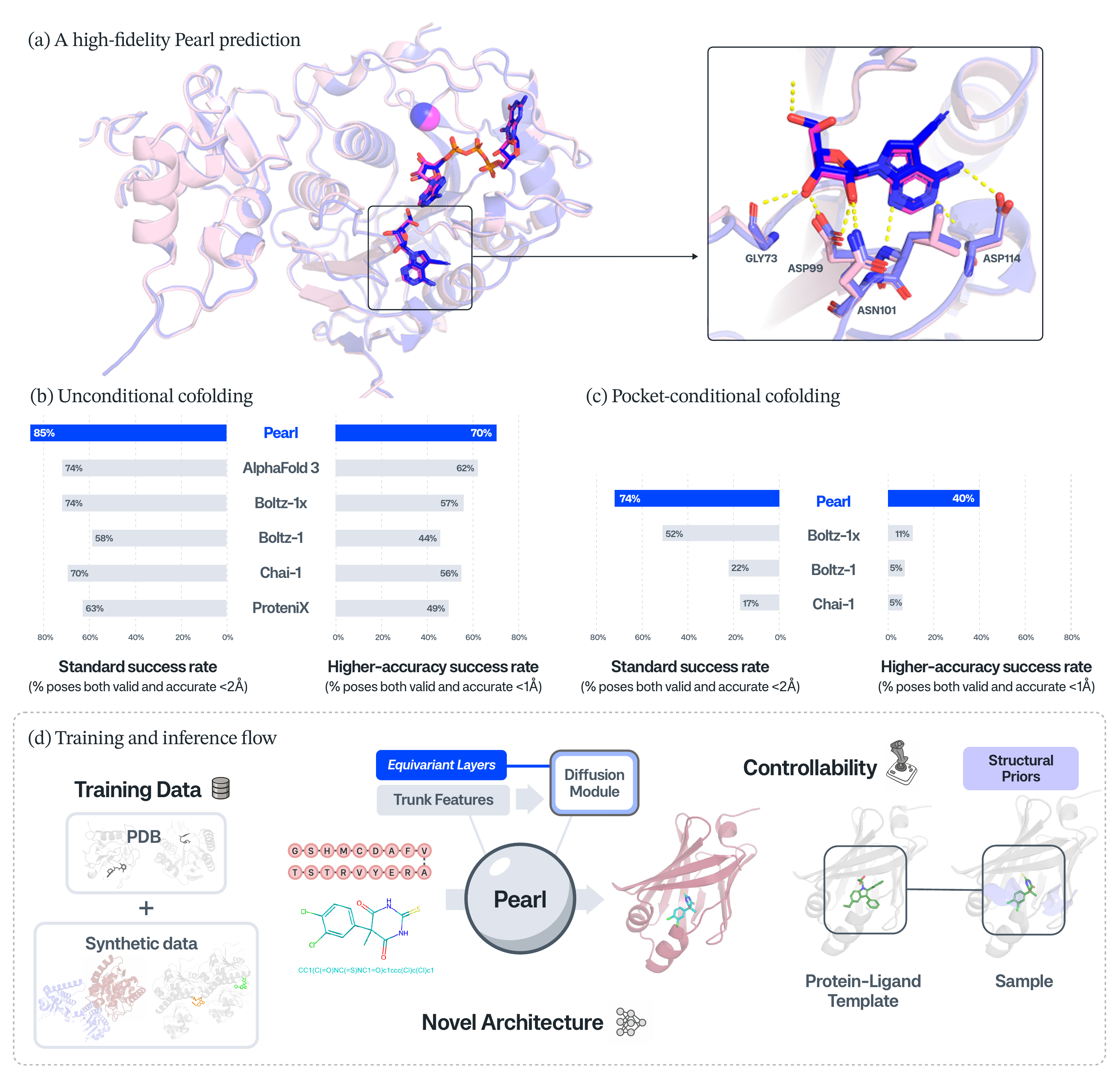}
    \caption{%
        \small
        (a) \pearl prediction (magenta) superimposed on the experimental ground truth (blue) for SARS-CoV-2 (PDB: 8S8X, in the \runsnposes, released March 2024); \pearl performance in the (b) unconditional (\runsnposes) and (c) pocket-conditional (\internalxtals) cofolding modes; (d) \pearl training and inference flow.
    }
    \label{fig:cover-image}
\end{figure}

Small molecule therapeutics are an essential component of modern medicine, but their discovery remains a slow, expensive, and high-risk endeavor.
A common approach involves building \textit{structure-activity relationship} (SAR) models to predict how changes to a ligand's chemical structure will affect its activity.
While both qualitative and quantitative SAR models have been successfully used to design drugs, they require large amounts of expensive experimental data, making the process inefficient and often yielding an incomplete understanding of the activity landscape, especially activity cliffs.
A key strategy to improve SAR modeling and to accelerate this process is \textit{structure-based drug design} (SBDD), where scientists design new ligands by applying chemical and physical principles to 3D protein-ligand structures.
Historically, SBDD has been limited by its reliance on experimentally determined structures, which are still slow and expensive to acquire.
The rise of computational methods presents an opportunity to apply this approach at an unprecedented scale, using predicted 3D structures---referred to as \emph{poses}---to triage candidates and to produce more information-rich SAR models.

Traditionally, 3D structure prediction has been performed using physics-based ligand docking methods~\cite{rarey_fast_1996,surflex,mcgann_gaussian_2003,gold,friesner_glide_2004,corbeil_variability_2012,venkatraman_flexible_2012,AutoDockVina}.
However, these methods tend to treat the protein (semi-)rigidly, and are unable to account for protein conformational changes that may occur during ligand binding~\cite{inducedfit1,inducedfit2,miller2021inducedfitdocking}.
On the other hand, early protein folding models treat the protein as fully flexible, but are unable to jointly predict protein-ligand complexes~\cite{af2}.
Recently, deep learning models have evolved beyond protein-only folding to predict more general biomolecular complexes.
Pioneering systems, such as \alphafold3 \cite{abramson2024accurate} and RoseTTAFold All-Atom~\cite{krishna2024rfaa}, have demonstrated that a single model could jointly model proteins, ligands, nucleic acids, and other cofactors, predicting structures directly from their sequences and chemical topologies.
The development of these models, and their descendants \cite{wohlwend2024boltz1, passaro2025boltz2, chai2024chai, bytedance2025protenix, liu2024helixfold3,qiao2024neuralplexer3}, is a major advancement in biomolecular modeling.
Crucially, these \emph{cofolding} models adapt the protein binding pocket geometry to different ligands, accounting for flexibility more fundamentally than previous approaches.
Despite this, achieving the reliability needed to advance drug discovery campaigns remains a significant challenge.

Reliable pose prediction faces three primary challenges. First, cofolding models learn from a relatively limited and biased corpus of experimentally determined structures. The Protein Data Bank (PDB)~\cite{berman2000protein} offers orders of magnitude less data than domains like text or images, and this corpus is skewed toward certain targets and chemotypes due to varying experimental difficulty and scientific interest~\cite{wwpdb_stats_2025}. This scarcity and bias limit generalizability, leading models to memorize common structures, rather than learn transferable rules~\cite{vskrinjar2025have}. Second, a useful binding pose must satisfy complex physical requirements, such as low ligand strain, shape complementarity, and favorable non-covalent interactions. Subtle violations can produce superficially plausible but physically incorrect \textit{"hallucinations"} that are not representative of real-world protein-ligand binding behavior. Finally, many existing models offer limited inference-time controllability, preventing them from using auxiliary structural information, such as a homologous ligand-bound structure, which are often available in real-world discovery programs.

To overcome these challenges, we introduce \pearl (\underline{P}lacing \underline{E}very \underline{A}tom in the \underline{R}ight \underline{L}ocation), a generative foundation model for biomolecular structure prediction (Figure~\ref{fig:cover-image}). 
\pearl addresses data scarcity and bias by training on a diverse mixture of experimental and synthetically generated protein-ligand complexes.
Its novel architecture and training protocol (Figure~\ref{fig:cover-image}) are designed to better learn the physical principles of binding, improving both generalization and the physical validity of generated poses.
Furthermore, \pearl is explicitly built for adaptability and use during drug discovery campaigns. 
Its novel multi-chain templating system acts as a flexible conditioning mechanism, allowing scientists to leverage auxiliary structural information about the target protein, cofactors, and related ligands.

This paper makes three key contributions.
First, we provide evidence for model performance scaling by training with large-scale synthetic data.
Second, we introduce key modeling innovations, including curriculum training and an architecture that encompasses an SO(3)-equivariant diffusion module.
Third, we develop a novel multi-chain templating system that incorporates non-polymeric information. This provides two key benefits: it enables conditioning on structural information during inference, and it supplies contextual data during training.
Taken together, these contributions establish \pearl as the new state-of-the-art model for protein-ligand cofolding.
\pearl achieves 85.2\% and 84.7\% success rates for generating accurate and physically valid poses on \runsnposes and \posebusters benchmarks, respectively---both significant (\pval{0.001}) improvements over \alphafold3 and all other baselines (\rmsdtwo~\AA and PB-valid). Notably, this high success rate shows almost no drop when physical validity checks are applied (a mere 0.7\% on \runsnposes and 0.4\% on \posebusters), highlighting \pearl's ability to generate almost exclusively physically plausible poses.
\pearl's performance advantage grows at the stricter, high-accuracy thresholds (\rmsdonea and PB-valid) that are essential for guiding medicinal chemistry efforts, where it exhibits an almost $4\times$ relative improvement over other models on a set of challenging, real-world targets (\internalxtals).

In the following sections, we contextualize \pearl in relation to prior work (Section~\ref{sec:related-work}), provide methodological detail---including its data, architecture, and inference strategies (Section~\ref{sec:methods})---present a comprehensive performance evaluation (Section~\ref{sec:results}), discuss the implications of our findings (Section \ref{sec:discussion}), and offer concluding remarks (Section \ref{sec:conclusion}).
This work has a dual aim: first to advance the development of foundation models for drug discovery, providing a path for cofolding systems that are not only state-of-the-art but also robustly generalizable, physically valid, and \emph{practically useful} for accelerating therapeutic design.
Second, and more broadly, we intend to push the frontier of generative AI for complex, physics-based problems in a low data regime.
We believe the principles and methods outlined here will be of interest to the broader scientific community at the intersection of machine learning, chemistry, and biology.

\section{Related Work}
\label{sec:related-work}

The computational prediction of protein-ligand binding poses has long been a cornerstone of SBDD, and it was traditionally carried out by physics-based ligand docking \cite{docking-review} (\eg,
FlexX~\cite{rarey_fast_1996},
Surflex-Dock~\cite{surflex},
FRED~\cite{mcgann_gaussian_2003},
GOLD~\cite{gold},
Glide~\cite{friesner_glide_2004},
MOE~DOCK~\cite{corbeil_variability_2012},
DOCK~\cite{venkatraman_flexible_2012},
Autodock Vina~\cite{AutoDockVina}).
These methods, which treat the pocket as (semi-)rigid, are either unable to treat induced fit---the protein conformational changes that may occur upon binding~\cite{inducedfit1,inducedfit2,miller2021inducedfitdocking}---or can only account for minor, local rearrangements of the binding pocket.
While there exist deep learning approaches for molecular docking \cite{corso2022diffdock,lee_genmol_2025}, including diffusion-based generative models, they do not jointly fold the protein and ligand, nor do they outperform classical docking methods~\cite{buttenschoen2024posebusters}.

A paradigm shift occurred with the advent of end-to-end folding models, most notably when \alphafold2 achieved near-experimental accuracy on the CASP14 challenge on protein monomers, starting from sequence and evolutionary information alone~\cite{af2,casp14}.
RoseTTAFold extended this approach to complexes with multiple protein chains~\cite{rosettafold2021}.
\alphafold3 (AF3) \cite{abramson2024accurate} generalized protein folding models to nearly all molecule types in the Protein Data Bank (PDB)~\cite{wwpdb_stats_2025}, inspiring a new generation of \emph{cofolding} models, such as
\chai1~\cite{chai2024chai},
\boltz1(x)~\cite{wohlwend2024boltz1},
\boltz2~\cite{passaro2025boltz2},
HelixFold3~\cite{liu2024helixfold3},
NeuralPlexer3~\cite{qiao2024neuralplexer3}, and
\protenix~\cite{bytedance2025protenix}.
Modern cofolding architectures typically share a blueprint, consisting of a trunk module (\eg, AF3's Pairformer) and a generative structure module, which are often denoising diffusion probabilistic models~\cite{diffusion}.
While \pearl is inspired by these approaches, it also altogether reflects a novel approach to cofolding.
\pearl incorporates geometric deep learning principles \cite{bronstein2017geometric} among other architectural, pretraining, and training improvements.
These innovations, together with \textbackslash{}pearl's unique inference-time controllability, enable state-of-the-art usability for practical small molecule drug discovery, including an enhanced ability to exploit known binding pockets and prior liganded structures.

A persistent bottleneck for all cofolding systems is their reliance upon relatively limited and biased training data from the PDB~\cite{burley2025updated}, which can lead to poor generalization to novel structures, instead relying on \emph{memorization} \cite{vskrinjar2025have}.
To mitigate this challenge, models have incorporated strategies like training on ``distillation data''~\cite{Melnyk2025} from other models \cite{ahdritz2024openfold} or integrating molecular dynamics and biochemical assay data~\cite{passaro2025boltz2}.
In contrast, \pearl's training corpus is augmented with a large-scale, dataset of synthetically generated protein-ligand complexes, exposing the model to wider chemical diversity.

Finally, controllability, \eg, the ability of users to incorporate structural priors during inference via templates, can improve the practical utility of cofolding models.
AF3 established per-protein-chain templates as the standard for cofolding models.
While scalable, this strategy can miss cross-chain interactions~\cite{abramson2024accurate} and more sophisticated approaches have emerged.
For example, later open source models introduced external experimental restraints and multimeric templates for protein complexes~\cite{chai2024chai,passaro2025boltz2}.
\pearl further generalizes multi-chain templates to additionally encompass non-polymeric components, such as cofactors and related ligands.

\section{Methods}
\label{sec:methods}

\pearl is a generative foundation model for protein-ligand structure prediction. This section outlines its key novel components and their functions.

\begin{figure}[t]
    \centering
    \includegraphics[width=0.95\linewidth]{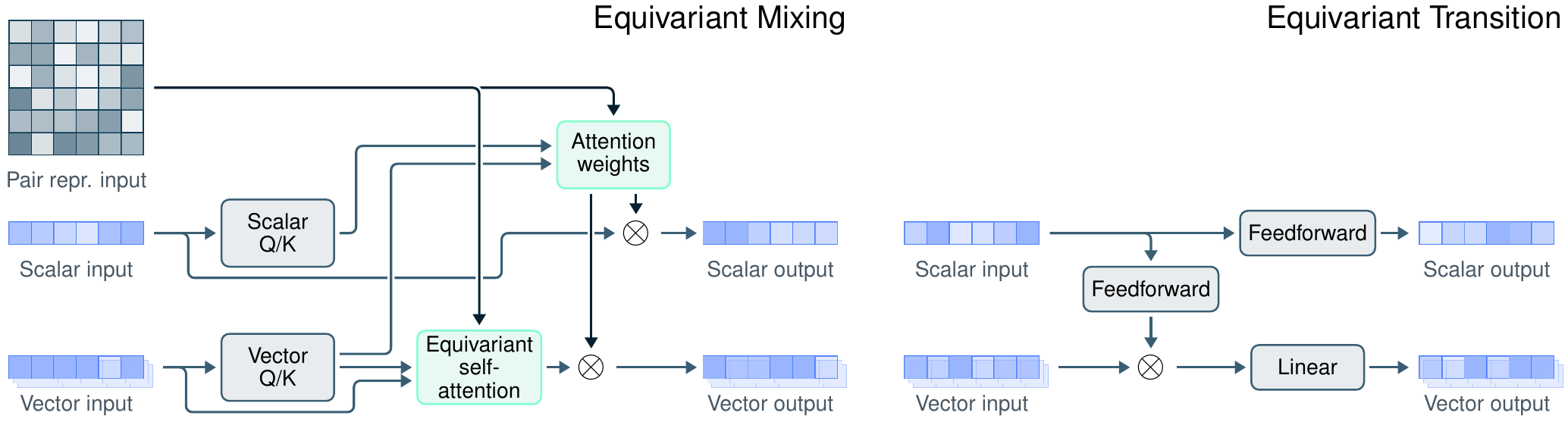}
    \caption{%
        Key components of the equivariant diffusion module, including the equivariant transformer architecture (left) and the equivariant feed-forward layer with a gated nonlinearity for vector components (right).
    }
    \label{fig:eqt}
\end{figure}

\paragraph{\pearl architecture}
\pearl's architecture consists of rotationally- and translationally-invariant representation modules (the ``trunk'') and diffusion modules responsible for predicting 3D coordinates.
The trunk, featuring a lightweight triangle multiplication (\texttt{trimul}) module for efficiency~\cite{pairmixer}, learns a rich, position-independent pairwise representation that conditions the diffusion module.
This design amortizes the trunk's high computational cost over the diffusion process's many sampling steps.
A combination of data augmentation and equivariant architecture achieves rotational and translational equivariance of the final 3D structure.
A key component is a novel $\text{SO}(3)$-equivariant diffusion module, which is unique among cofolding models. This module is constructed from equivariant transformer (EqT) blocks (Figure~\ref{fig:eqt}).
By obeying $\text{SO}(3)$ symmetry by construction, this module works in synergy with data augmentation to improve training sample efficiency and ensure predictions naturally respect the rotational symmetries of 3D space.

\paragraph{Templates}
To provide the model with a richer and more complete structural prior, \pearl generalizes the standard templating approach from protein-only templates to templates that also include non-polymeric components. The goal is to supply the model with a coherent, ``holo-like'' pocket environment---that is, a ligand-bound conformational state, in contrast to the unbound (apo) state. This provides context from similar ligands and their interactions with protein and cofactor components. In prospective settings, this templating form exposes crucial controllability and specificity for drug discovery scientists.

\paragraph{Training}
\pearl's novel training recipe enhances generalization and structural fidelity through its data mixture, curriculum, and optimization strategies. We train \pearl on a diverse dataset combining curated PDB structures, monomer distillation data (OpenFold \cite{ahdritz2024openfold} and the AlphaFold Database \cite{varadi2024alphafold}), and a \textit{novel, large-scale synthetic dataset}. Derived from public data, this dataset is generated using physics-based methods with diverse virtual ligands. The synthetic data introduces important chemical diversity beyond experimental data limitations.

Training follows a \textbf{five-stage curriculum} that progressively increases task complexity and data diversity. Initial stages use smaller crop sizes and simpler data mixtures (non-templated PDB, monomer distillation). Later stages gradually introduce more complex structural priors and templating information across datasets. The training applies structural templates across the datasets to vary the structural context.

To manage computational cost, \pearl employs a \textbf{conservative \texttt{bfloat16} (\texttt{bf16}) mixed-precision strategy}. Computationally intensive trunk operations (\eg, triangle ops via NVIDIA \cuequivariance \cite{nvidia_cuequivariance_github}, \layernorm \cite{ba2016layer}) use \texttt{bf16}. However, numerically sensitive components (losses, coordinate projections), unstable operations (\eg, softmax), and all model weights remain in full \texttt{fp32} precision to ensure stability. This balanced approach significantly improves efficiency without compromising numerical stability.

\paragraph{Inference-time strategies}
\pearl uses several in-context learning strategies for high-quality pose generation, ranging from two fundamental operating modes to advanced techniques for fine-grained control. The dual retrieval capabilities of its versatile multi-chain templating system and pocket conditioning enable \pearl to generate structures in two distinct cofolding modes tailored to different use cases: \textit{unconditional} and \textit{(pocket-aware) conditional}.
In the unconditional cofolding mode, \pearl predicts the complex structure using only the protein's amino acid sequence and the ligand's topology. The method may use this sequential information to search for template structures and evolutionary information (\ie MSAs) that the model may also use, similarly to AF3. The unconditional mode is useful for novel targets where the binding pocket is not known \emph{a priori}.
The conditional cofolding mode enables a common drug discovery scenario in which a reference apo or holo structure, or a hypothesized binding pocket, is available to guide generation.

Optional guidance and steering techniques can further control \pearl's sampling by modifying the denoising trajectory to enforce specific constraints or physical priors with a configurable parameter governing the strength of the enforcement. These methods are developed for practical drug discovery needs and allow for the additional use of a scientist's contextual knowledge without requiring costly model retraining. \pearl also offers the ability to modulate pose diversity at inference time.

\section{Results}
\label{sec:results}
This section comprehensively evaluates \pearl against several baselines on public and proprietary benchmarks, addressing performance, practical utility in drug discovery, and the factors underlying its capabilities. 

\subsection{Evaluation methodology}
\label{sec:results:eval}

All training and evaluations are conducted on a high-performance computing cluster with NVIDIA H100 and H200 GPUs. We use optimized kernels from \cuequivariance (v0.6.0) \cite{nvidia_cuequivariance_github} and a custom CUDA \layernorm kernel~\cite{bytedance2025protenix} to accelerate key trunk operations. 
We evaluate \pearl against the following baselines: \alphafold3~\cite{abramson2024accurate}, \boltz1(x)~\cite{wohlwend2024boltz1}, \boltz2~\cite{passaro2025boltz2}, \chai1~\cite{chai2024chai}, and \protenix~\cite{bytedance2025protenix} on diverse benchmarks. These benchmarks include the public \runsnposes~\cite{vskrinjar2025have} (RnP) and \posebusters~\cite{buttenschoen2024posebusters} datasets, as well as a proprietary \internalxtals dataset. \internalxtals dataset consists of 111 structures from a variety of internal programs.
It is designed to test real-world generalization. To ensure consistent benchmarking, the public RnP and \posebusters datasets were subsetted to structures released after \fmtdate{2023-06-01} and \fmtdate{2021-10-01}, respectively (see Appendix~\ref{app:eval:datasets}).
\pearl's training set consisted strictly of publicly available structures from the PDB released on or before \fmtdate{2021-09-30}; synthetic complexes were derived only from the pre-\fmtdate{2021-09-30} publicly available structures.
The \pearl model evaluated in this paper was trained without any proprietary experimental data.

We compare models in both unconditional cofolding and, when available, (pocket-aware) conditional cofolding modes to test for controllability of generation with structural priors (Appendix~\ref{app:eval:regimes}).
The primary accuracy metric is ligand root mean squared deviation (RMSD)~\cite{gilson2025assessment}.
We use a thresholded RMSD success rate (\lttwoa or \ltonea denotes success), supplemented by physical plausibility checks from the \posebusters~\cite{buttenschoen2024posebusters} validation suite (PB-valid; Appendix~\ref{app:eval:metrics}).
To decouple the evaluation of \textit{pose generation} from \textit{pose selection}, we adopt the \bestk protocol common in generative modeling~\cite{chen2021evaluating}, evaluating for $k=1, 5,~\mathrm{and}~20$ out of 20 samples (see Appendix~\ref{app:experimental-setup:bestk} for discussion).
We report the main results for the \best5 protocol, common in the generative modeling literature; the complete set of results is in Appendix~\ref{app:metrics}.

\begin{figure}[t!]
    \centering
    \vspace{-1ex}
    
    \begin{subfigure}[b]{0.9\linewidth}
        \includegraphics[width=1.\linewidth]{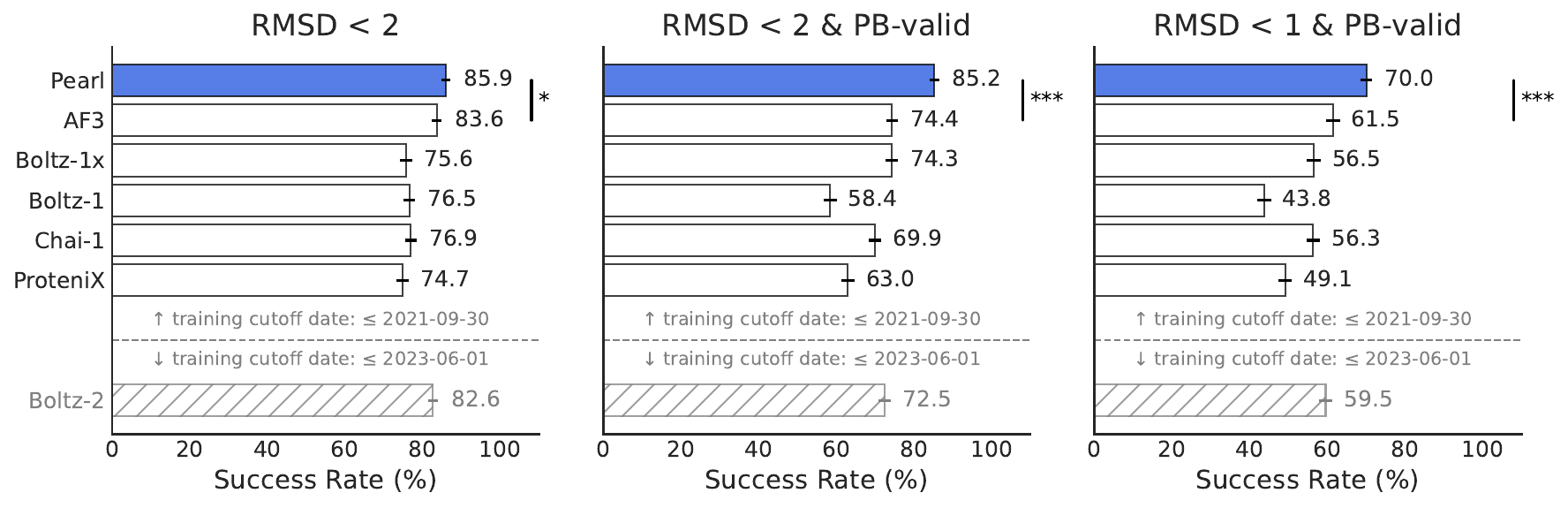}
        \caption{\runsnposes benchmark.}
        \label{fig:results:rnp-rmsd-best@5-cofolding}
    \end{subfigure}\\[1ex]

    \begin{subfigure}[b]{0.9\linewidth}
        \includegraphics[width=1.0\linewidth]{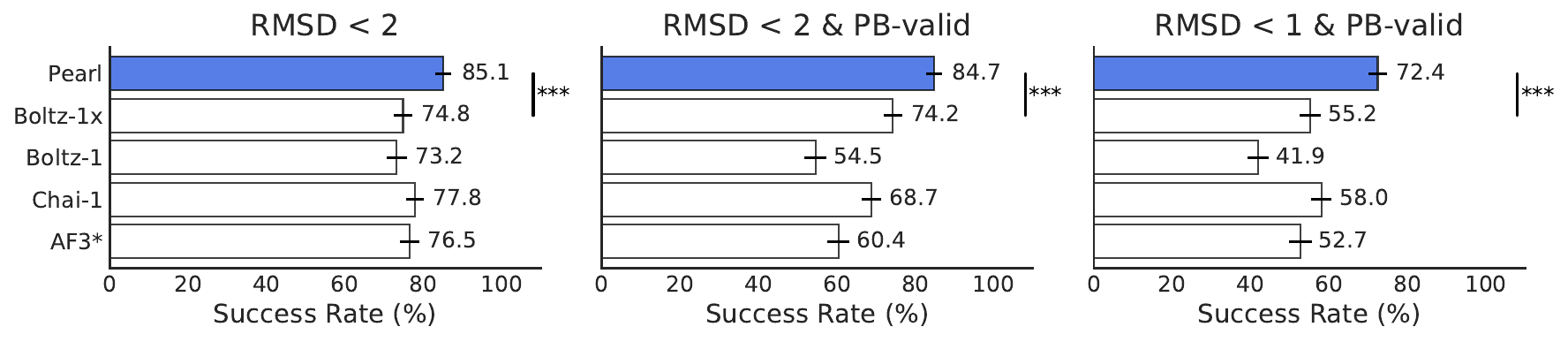}
        \caption{\posebusters benchmark.}
        \label{fig:results:pb-rmsd-best@5-cofolding}
    \end{subfigure}\\[1ex]
    
    \begin{subfigure}[b]{0.9\linewidth}
        \includegraphics[width=1.0\linewidth]{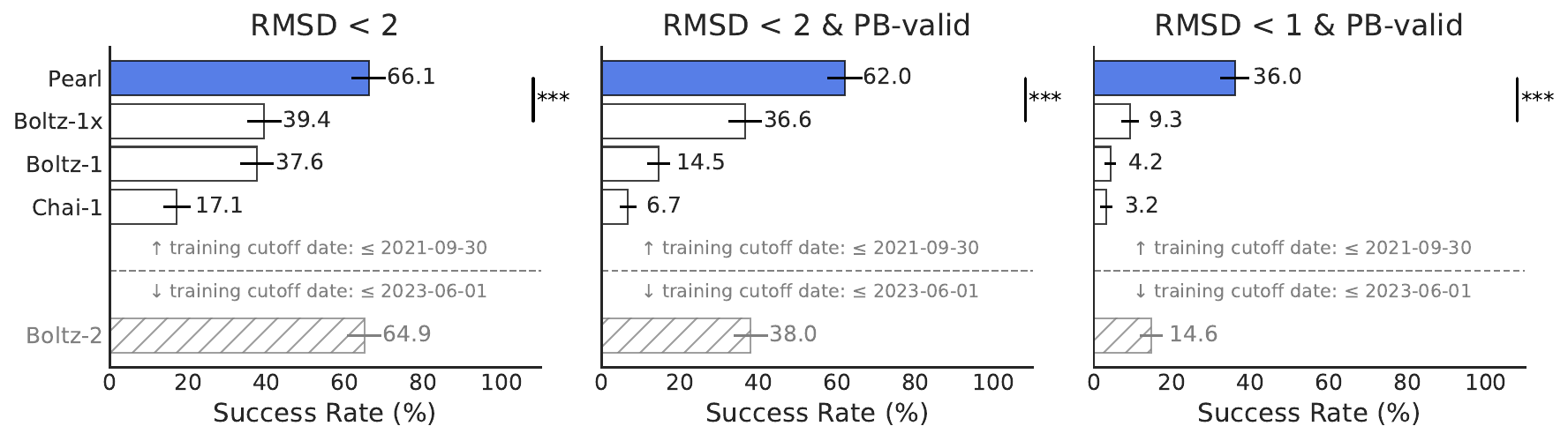}
        \caption{\internalxtals benchmark.}
        \label{fig:results:ix-rmsd-best@5-cofolding} 
    \end{subfigure}

    \caption{
        \small
        \textbf{\pearl demonstrates state-of-the-art performance in the unconditional cofolding mode across multiple benchmarks.}
        Shown are \best5 success rates for generating accurate (\rmsdonea\ and \lttwoa) and physically valid (PB-valid) poses.
        Dashed lines group models for fair comparison (\eg, based on the training cutoff dates).
        Note~1(*): \alphafold3 results for \posebusters use officially released metrics for poses selected for max confidence out of 25 samples (not \best5).
        Note~2: \alphafold3 was excluded from the \internalxtals benchmark due to license restrictions.
        Note~3: Trained with additional public data up to \fmtdate{2023-06-01}, \boltz2 is not directly comparable to the other evaluated models, which use \fmtdate{2021-09-30} or earlier training cutoff.
        Note~4: Statistical significance (one-sided t-test) is shown as: $*: \mathrm{p\leq0.05}$, $**: \mathrm{p\leq0.01}$, $***: \mathrm{p\leq0.001}$.
    }
    \label{fig:public}
\end{figure}

\subsection{\pearl is a state-of-the-art cofolding model}
\label{sec:results:cofolding}

Benchmarked against leading cofolding models on public and proprietary datasets, \pearl consistently establishes a new state-of-the-art, delivering significantly improved success rates for accurate and physically valid poses compared to all baselines. The following sections detail these results, focusing on public benchmarks (Section \ref{sec:public}) and \pearl's generalization capabilities (Section \ref{sec:generalization}).

\subsubsection{\pearl achieves state-of-the-art performance on public benchmarks}
\label{sec:public}

Figure~\ref{fig:public} shows \pearl's significant and consistent improvement over \alphafold3 and other methods across progressively stricter metrics. For the standard \rmsdtwoa metric (\best5), \pearl attains top-ranking success rates of 85.9\% on \runsnposes and 85.1\% on \posebusters (Figures~\ref{fig:results:rnp-rmsd-best@5-cofolding} and \ref{fig:results:pb-rmsd-best@5-cofolding}, left). This advantage grows when also evaluating for physical validity (\rmsdtwoa and PB-valid, \best5). On \runsnposes, \pearl's 85.2\% success rate is a 14.5\% relative improvement over the next best models, \alphafold3 (74.4\%) and \boltz1x (74.3\%) (Figure~\ref{fig:results:rnp-rmsd-best@5-cofolding}, middle). Similarly, on \posebusters, its 84.7\% success rate represents a 14.2\% relative improvement over \boltz1x (74.2\%) and a 40\% relative gain over \alphafold3's officially released max confidence result (60.4\%) (Figure~\ref{fig:results:pb-rmsd-best@5-cofolding}, middle).
On \posebusters, \pearl in the \best{1} regime (selecting 1 pose at random) still maintains an 8\% relative advantage over \alphafold{3}, which selects one max confidence pose from over 25 generated poses.
In the \posebusters \best{20} setting \pearl has an 46\% relative advantage over \alphafold{3}'s max-confidence. For more details, see Appendix~\ref{app:metrics}.

Notably, \pearl's success rate shows almost no drop when validity checks are applied---a mere 0.7\% drop on \runsnposes and 0.4\% on \posebusters (Figures ~\ref{fig:results:rnp-rmsd-best@5-cofolding} and ~\ref{fig:results:pb-rmsd-best@5-cofolding}, middle)---highlighting its ability to generate almost exclusively physically plausible poses.
A comprehensive breakdown of metrics at different aggregation levels (\best1, \best5, \best{20}) is provided in Appendix~\ref{app:metrics} (Tables~\ref{tbl:best1_metrics}, \ref{tbl:best5_metrics}, \ref{tbl:best20_metrics}).

\subsubsection{\pearl demonstrates strong generalization to novel drug targets}
\label{sec:generalization}

Beyond public benchmarks, we evaluate \pearl's generalization on the \internalxtals dataset, designed to test performance on targets and ligands structurally and chemically dissimilar to the public training corpus (Figure~\ref{fig:results:ix-rmsd-best@5-cofolding}).
In the unconditional cofolding mode, \pearl achieves a 66.1\% success rate for \rmsdtwoa (\best5), a 67.8\% relative improvement over the next best open source model (\boltz1x at 39.4\%).
For physically valid poses (\rmsdtwoa and PB-valid, \best5), \pearl maintains a strong lead with 62.0\% success, a 69.4\% relative gain over \boltz1x (36.6\%) (Figure~\ref{fig:results:ix-rmsd-best@5-cofolding}, left and middle).
This robust performance on challenging, real-world data demonstrates that \pearl's advantages extend beyond academic benchmarks.

\begin{figure}[t!]
    \centering
    \includegraphics[width=1.0\linewidth]{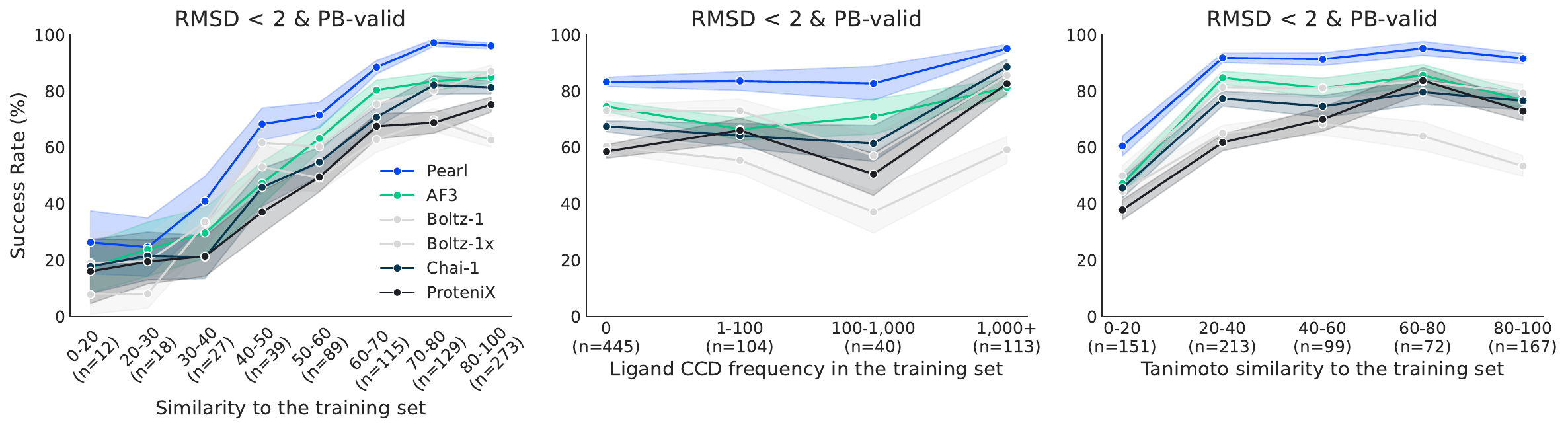}
    \caption{%
        \small
        Stratified analysis of \pearl's generalization on the \rnp in the unconditional cofolding mode.
        Shown are the \best5 success rate for generating valid poses (\rmsdtwoa and PB-valid) when stratified by (left) overall similarity to the training set (product of binding pocket coverage and combined overlap score [SuCOS] of the ligand pose \cite{vskrinjar2025have}), (middle) ligand frequency, and (right) Tanimoto similarity. \pearl exhibits strong generalization in the most challenging, low-similarity regimes, where it generally leads other models. The number of test examples in each slice is denoted by $n$. \boltz2 is excluded as its expanded training set does not conform to the same similarity measures.
    }
    \label{fig:results:rnp-rmsd-stratified}
\end{figure}

To further probe generalization, Figure~\ref{fig:results:rnp-rmsd-stratified} analyzes \pearl's performance on the \rnp benchmark stratified by similarity to the training data, following~\cite{vskrinjar2025have}.
Focusing on the lowest similarity buckets---the most rigorous test of generalization by this metric---reveals \pearl's consistent advantage.
\pearl leads when generalizing to novel protein pockets (similarity to the training set < 0.2, as defined in \cite{leung_sucos_2019,vskrinjar2025have}), novel ligands (frequency = 0), and dissimilar chemotypes (Tanimoto similarity < 0.2).
This strong performance across different generalization axes indicates \pearl learns transferable rules rather than memorizing training examples, establishing it as a suitable foundation model for diverse drug discovery tasks.

\subsection{\pearl's predictions are relevant for drug discovery}

This section establishes \pearl's practical relevance for drug discovery. Section \ref{sec:highres} demonstrates its superior performance at the high-accuracy thresholds required for fine-grained molecular design.
We also showcase its lead in accuracy and controllability in the presence of structural priors in the conditional cofolding mode (Section \ref{sec:pocket}). Together, these results show that \pearl generates poses that are not just statistically accurate but better poised to be genuinely useful for guiding real-world therapeutic design.

\begin{figure}[t!]
    \centering
    \begin{subfigure}[b]{0.9\linewidth}
        \includegraphics[width=1.0\linewidth]{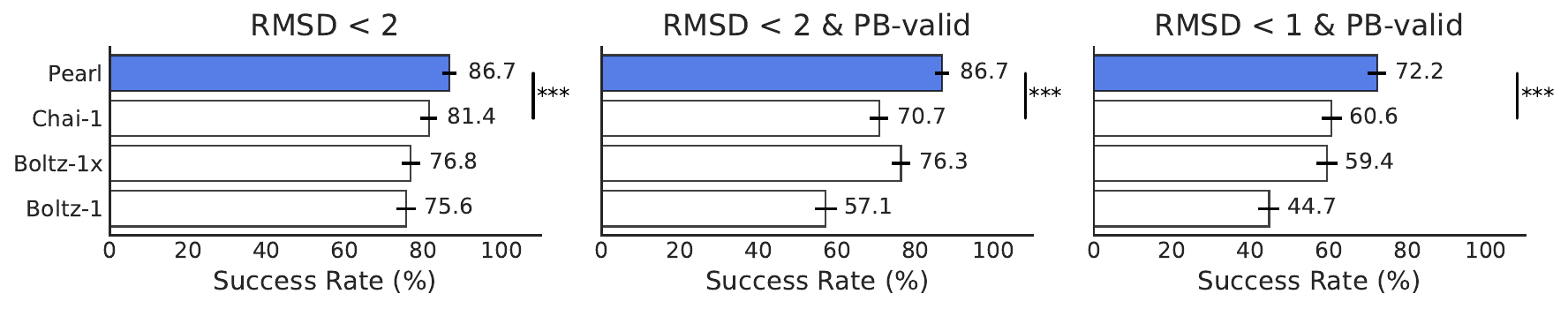}
        \caption{\posebusters benchmark.}
        \label{fig:results:pb-rmsd-best@5-cond}
    \end{subfigure}\\[1ex]
    
    \begin{subfigure}[]{0.9\linewidth}
        \includegraphics[width=1.0\linewidth]{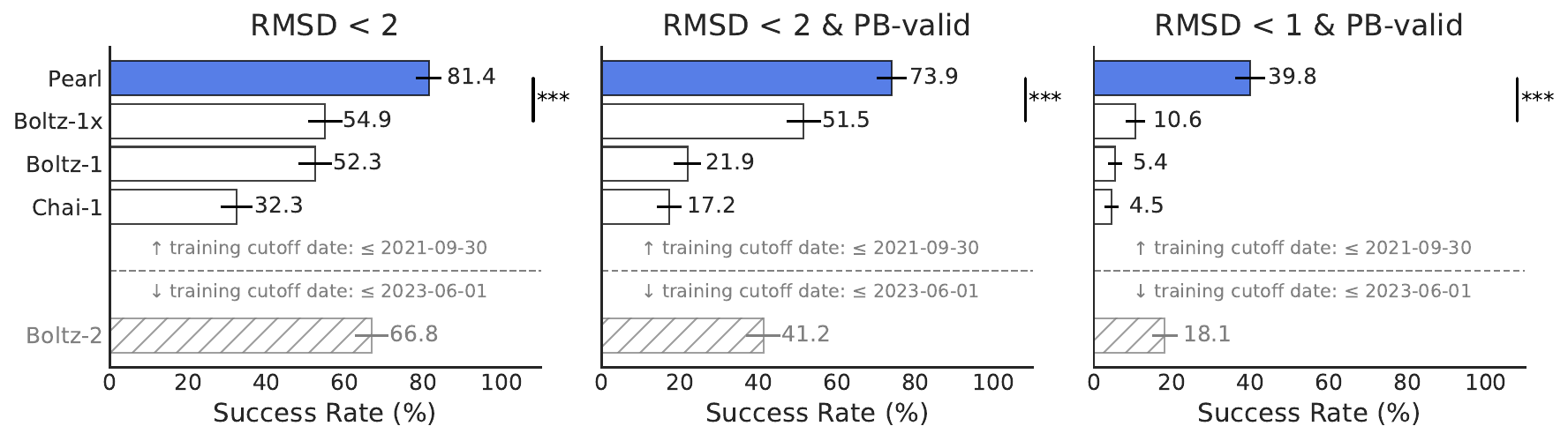}
        \caption{\internalxtals benchmark.}
        \label{fig:results:xrtl-rmsd-best@5-cond}
    \end{subfigure}

    \caption{%
        \small
        \textbf{\pearl demonstrates state-of-the-art performance in the conditional cofolding mode across multiple benchmarks.} 
        Metrics shown are \best5 (from 20 samples).
        Dashed lines group models for fair comparision based on the training cutoff dates.
        Note 1: \alphafold3 was excluded due to license restrictions.
        Note 2: Trained with data up to \fmtdate{2023-06-01}, \boltz2 is not directly comparable to models using earlier cutoffs.
    }
    \label{fig:results:conditional-benchmarks}
\end{figure}

\subsubsection{\pearl excels at the high-accuracy thresholds required for drug discovery}
\label{sec:highres}

While \rmsdtwoa is a standard benchmark metric, guiding fine-grained molecular design for affinity improvements often requires higher accuracy.
We therefore evaluate models with a stricter threshold of \rmsdonea and PB-valid (\best5).
At this high-accuracy threshold, \pearl's performance advantage becomes even more pronounced across all benchmarks (Figure~\ref{fig:public}, right panels).

On public benchmarks, \pearl reaches success rates of 70.0\% on \runsnposes and 72.4\% on \posebusters, uniquely maintaining over 70\% success while baselines' performance drops significantly (Figures~\ref{fig:results:rnp-rmsd-best@5-cofolding} and \ref{fig:results:pb-rmsd-best@5-cofolding} right). For instance, on \runsnposes, the next best model, \alphafold3, achieves 61.5\%.
On \posebusters, the next best comparable model (\chai1) achieves 58.0\%.
This state-of-the-art high-resolution performance is confirmed on the \internalxtals benchmark (Figure~\ref{fig:results:ix-rmsd-best@5-cofolding}, right).
Here, \pearl achieves a 36.0\% success rate at \rmsdonea and PB-valid. This represents a nearly 4-fold relative improvement over the strongest comparable baseline (\boltz1x at 9.3\%) and more than double the success rate of \boltz2 (14.6\%), which benefited from a later training data cutoff. This demonstrates \pearl's superior ability to generate the high-fidelity poses critical for medicinal chemistry.

\subsubsection{\pearl's structure-aware conditioning provides high-fidelity poses for guided research}
\label{sec:pocket}

To evaluate \pearl's utility for practical structure-based drug design, we assessed its performance in the (pocket-aware) conditional cofolding mode, where structural priors (\eg, known apo/holo structures) guide pose generation.
In this mode, \pearl consistently outperforms all comparable baselines, demonstrating its effectiveness as a controllable tool (Figure~\ref{fig:results:conditional-benchmarks}).
On the public \posebusters benchmark (Figure~\ref{fig:results:pb-rmsd-best@5-cond} middle and right), \pearl's success rate is 86.7\% (\rmsdtwoa and PB-valid, \best5), a 14\% relative improvement over \boltz1x (76.3\%).
At the stricter \rmsdonea and PB-valid threshold, its success rate of \textbf{72.2\%} is significantly ahead of \chai1 (60.6\%).
This guided approach delivers even stronger results on the \internalxtals benchmark (Figure~\ref{fig:results:xrtl-rmsd-best@5-cond}), with a 73.9\% success rate (\rmsdtwoa and PB-valid, \best5), a 43\% relative improvement over \boltz1x (51.5\%).
At the high-accuracy \rmsdonea threshold on this proprietary set, \pearl's \textbf{39.8\%} success rate is nearly double that of \boltz2 (18.1\%)---despite \boltz2 benefiting from a later training data cutoff---and almost $4\times$ better than \boltz1x (10.6\%).
All evaluated models are provided the same pocket residues for conditioning (see Appendix~\ref{app:eval:regimes}).
These results confirm \pearl's superior performance and controllability when leveraging structural priors common in drug discovery workflows.

\subsection{Key enablers of \pearl's performance}
\label{sec:ablation}

Having established \pearl's state-of-the-art performance, this section highlights two key enablers, presenting evidence for the contributions of large-scale synthetic data and the efficiency gains from architectural choices and hardware acceleration.

\begin{figure}[t]
    \centering
    \begin{minipage}[b]{0.48\textwidth}
        \includegraphics[width=\linewidth]{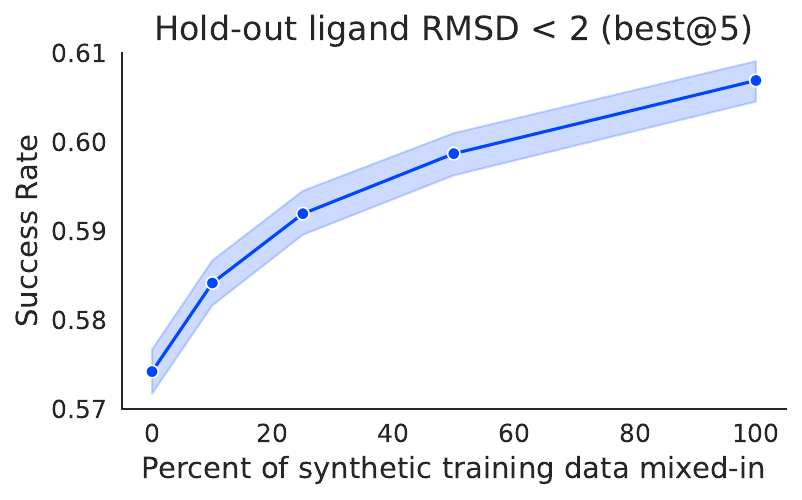}
    \end{minipage}
    \hfill
    \begin{minipage}[b]{0.48\textwidth}
        \small
        \begin{tabular}{@{}l|r@{}}
            \toprule
            Unique proteins & 910 \\
            Average number of ligands per protein & 640 \\
            Tanimoto similarity of ligands vs. reference & 0.364 \\ \midrule
            Total synthetic structures & 582,065 \\
            \bottomrule
        \end{tabular}
        \vspace{5.5ex}
    \end{minipage}
    \caption{%
        \small
        \textbf{\pearl's performance scales with the addition of synthetic data.} The plot (left) shows a monotonic improvement in success rate (\best5, \rmsdtwoa) for a smaller variant of \pearl on a fixed hold-out set as the proportion of synthetic data increases. The table (right) summarizes the scaling experiment's dataset, which included 582,065 synthetic structures across 910 proteins. A larger synthetic dataset was used to train the flagship \pearl model.
    }
    \label{fig:synth-data-scaling}
\end{figure}

\paragraph{Synthetic data improves generalization:}A core hypothesis in this work is that large-scale synthetic data can overcome PDB limitations and improve generalization. Our experiments with a smaller variant of \pearl validate this, revealing a clear scaling relationship: \pearl's success rate (\rmsdtwoa, \best5) on a fixed hold-out set increases monotonically with the proportion of synthetic training data  mixed into the training corpus (Figure~\ref{fig:synth-data-scaling}). This data scaling trend confirms that augmenting experimental structures with diverse synthetic examples is a driver of \pearl's robust performance and generalization.

\begin{figure}
    \centering
    \includegraphics[width=0.95\textwidth]{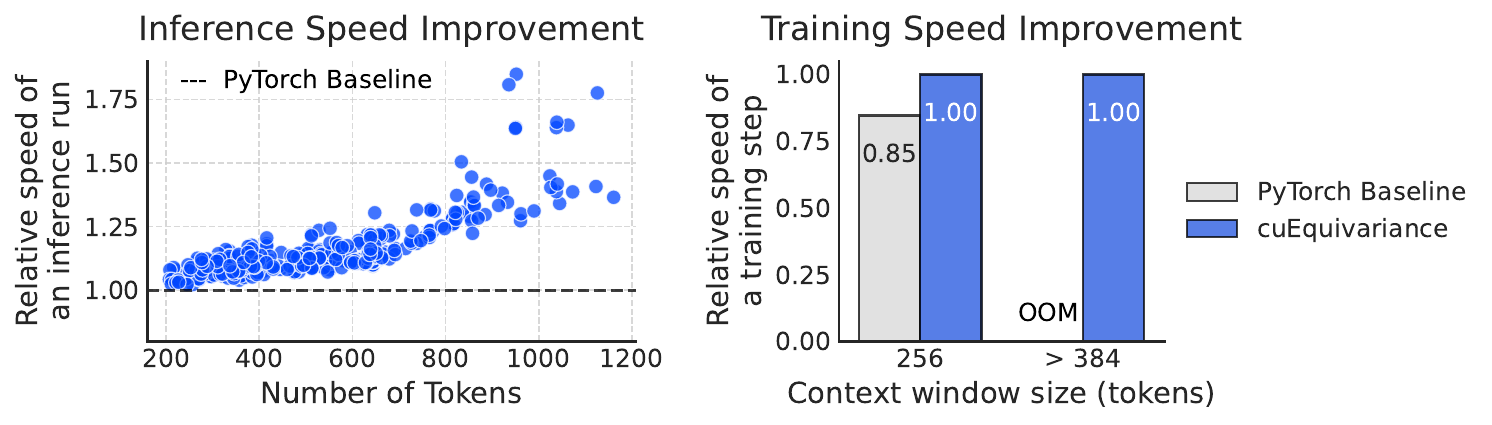}
    \caption{%
        \small
        NVIDIA's optimized \cuequivariance kernels provide significant acceleration for \pearl relative to a vanilla PyTorch baseline. The kernels deliver a 10--80\% speedup at inference time (left) and a 15\% training speedup (right).
    }
    \label{fig:cuequivariance}
\end{figure}

\paragraph{Architectural innovations and hardware acceleration improve efficiency:}
\pearl's architectural design significantly enhances computational efficiency.
The conservative \texttt{bfloat16} (\texttt{bf16}) mixed-precision strategy alone increases training speed by 22\% and reduces memory usage by 11\% compared to full \texttt{fp32} precision, without impacting final model accuracy.
Optimized kernels from NVIDIA's \cuequivariance (v0.6.0) \cite{nvidia_cuequivariance_github} library for computationally intensive operations in the model's trunk and diffusion modules yield further acceleration.
These kernels provide substantial speedups relative to a vanilla PyTorch baseline: an additional 15\% speedup during training and a 10--80\% speedup at inference time, with greater inference gains observed for larger inputs (more tokens) (Figure~\ref{fig:cuequivariance}).
Moreover, the GPU memory savings from \cuequivariance enable training with larger context window sizes.
These combined optimizations---mixed precision and hardware-accelerated kernels---were critical for executing the large-scale training and evaluation in this work.
Appendix~\ref{app:mixed-precision} provides more details on the mixed-precision implementation.

\subsection{Case studies}
\label{sec:results:case-studies}

\begin{figure}[t!]
    \centering
    \includegraphics[width=1.0\linewidth]{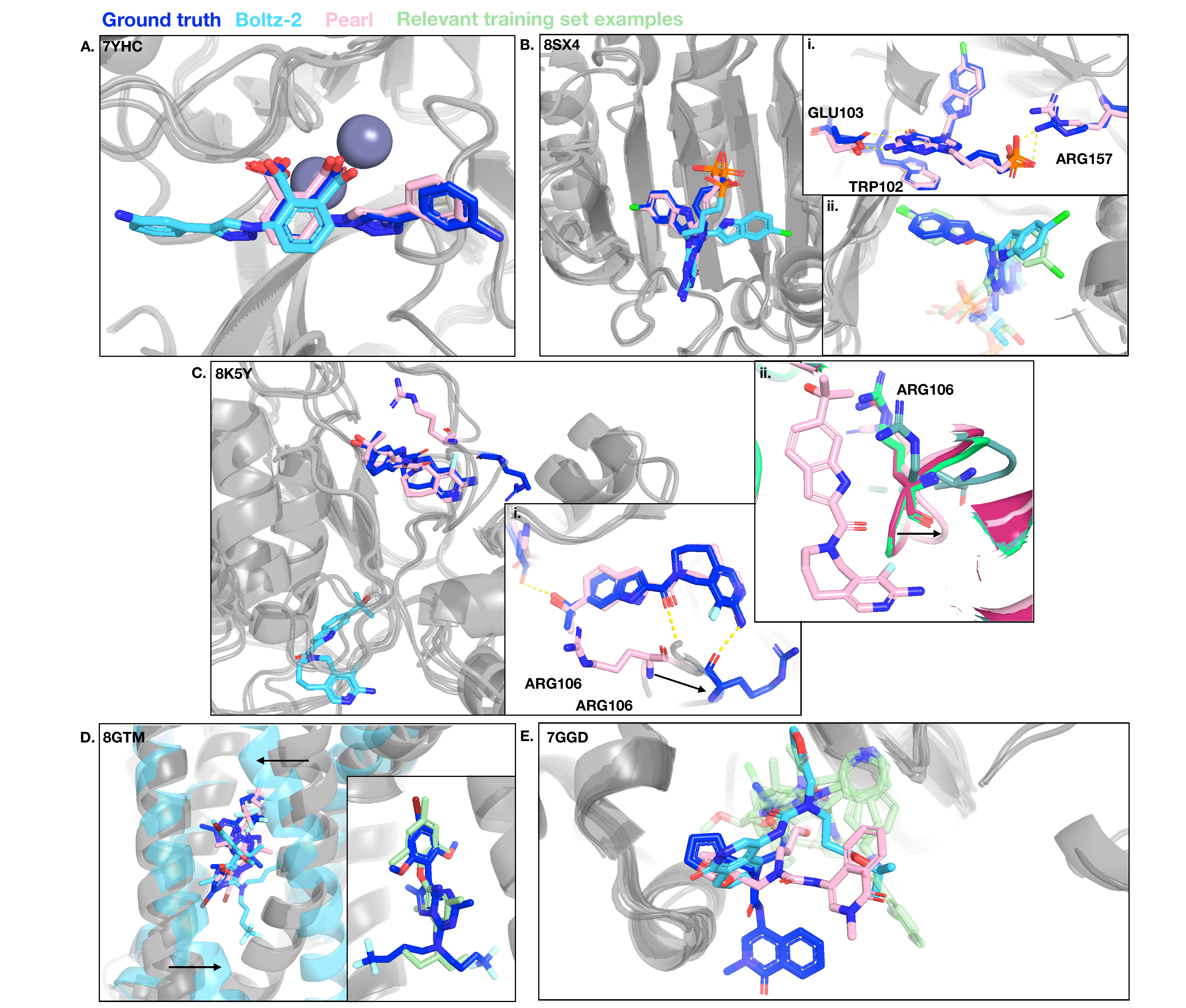}
    \caption{%
        \small
          \textbf{Qualitative analysis of \pearl's successes and common cofolding failure modes.} The ground truth is shown in dark blue, \pearl's prediction in pink, \boltz2's prediction in light blue, and relevant training set examples in green. \textit{\pearl Successes} (A, B, C): \pearl correctly predicts challenging poses where \boltz2 fails, such as for an inhibitor of VIM-2 MBL (A), an inhibitor of eIF4E (B), and a proMMP-9 inhibitor (C). \textit{Common Failure Modes} (D, E): Both models can place ligands in pockets that are more frequently observed in the training data, leading to incorrect predictions for inhibitors of CRF1R (D) and SARS-CoV-2 main protease (E).
    }
    \label{fig:results:rnp-qualitative}
\end{figure}

Complementing the quantitative benchmarks, we qualitatively analyze \pearl and \boltz2 poses drawn from \runsnposes in the unconditional cofolding mode.
The five case studies highlight common cofolding failure modes, including both those that \pearl overcomes and those that it still struggles to resolve (Figures~\ref{fig:results:rnp-qualitative}~and~\ref{fig:results:rnp-rmsd-qualitative}). 
Overall, this analysis links key failures to training set memorization.
Models often fail to predict novel pockets or conformational changes and exhibit a bias for frequently occupied pockets, even when binding to those pockets is incompatible with the predicted conformation.
This behavior, likely sensitive to training data composition and sampling, indicates a reliance on memorization over full generalization. We also show cases that underscore the necessity of strict \rmsdonea thresholds.
Many \rmsdtwoa poses miss critical interactions, limiting their practical utility for potency prediction or compound ideation in drug discovery.

\subsubsection{\pearl successes}
Figure~\ref{fig:results:rnp-qualitative}A shows \pearl correctly predicting the pose of a phthalic acid-based VIM-2 MBL inhibitor precursor.
\boltz2 identifies the correct overall pocket and key phthalic acid-Zn$^{2+}$ interactions, but it fails to place the aminophenyl and triazole rings in their correct subpocket.
This is despite both models seeing training examples occupying both sides of the pocket (\eg, 1JJT, 1KR3, 3WXC, 5LCF).
Similarly, for an inhibitor of the rate-limiting translation factor eIF4E  (Figure~\ref{fig:results:rnp-qualitative}B), \pearl correctly predicts the binding pocket and hydrogen bonds formed with Arg157, Glu103 and Trp102 (Figure~\ref{fig:results:rnp-qualitative}B i. inset).
\boltz2, however, misplaces the indole ring and the phosphate tail, placing the indole in an alternative subpocket that is seen in the training set (\eg, 5EHC, 6YLR).
This demonstrates a memorization failure, as data for the correct subpocket was also available in the training set (4DUM, Figure~\ref{fig:results:rnp-qualitative}B ii. inset).
Figure~\ref{fig:results:rnp-qualitative}C shows that \pearl correctly predicts a \lttwoa pose for a proMMP-9 catalytic domain inhibitor, while \boltz2 predicts the incorrect pocket. In fact, while all the baselines predict a > 2.0~\AA pose, only \pearl predicts a sub 1~\AA pose (0.96~\AA).

While \pearl is sufficiently accurate to be useful, it is not without imperfections.
For example, in the case of the proMMP-9 catalytic domain inhibitor, \pearl does not predict the correct loop conformation, and as a consequence the ligand fails to form an important loop backbone interaction.
This is likely a memorization artifact, as earlier structures (1GKD, 2OVZ, 2OVX, 6ESM) were crystallized without the loop domain.
Additionally, related proteins (proMMP-1, proMMP-7, and proMMP-9desFnII) lack the crucial interacting arginine in their homologous loop and that region is unresolved.
Overall, a similar loop conformation is sampled in only a few training structures (\eg, 5TH6, 5UE3, 5UE4), and superimposing this conformation clashes with the inhibitor (Figure~\ref{fig:results:rnp-qualitative}C ii. inset).
Nevertheless, \pearl is the only model that manages to predict a usefully accurate pose, even without capturing the exact conformation of the loop.

\subsubsection{Common failure modes}\vspace{-0.75ex}
Next, we examine cases where multiple cofolding models fail (Figure~\ref{fig:results:rnp-qualitative}D and \ref{fig:results:rnp-qualitative}E).
In the case of the CRF1R antagonist, BMK-C203, both \pearl and \boltz2 place the ligand in the right pocket, but they fail to predict the correct ligand conformation.
While \pearl does predict the correct pocket shape, \boltz2 predicts a more closed conformation of the pocket.
Interestingly, there are many structures of CRF1R and related proteins bound to ligands in other pockets (\eg, 3PDS, 4QKX, 4PHU, etc.), but only one where the ligand, CP-376395, occupies the same pocket as BMK-C203 (4Z9G, Figure~\ref{fig:results:rnp-qualitative}D inset).
It is possible that the abundance of structures with ligands in other pockets lead the cofolding models to struggle when placing ligands into the BML-C203 pocket. 
Another failure mode is exemplified by the SARS-CoV-2 main protease inhibitor, Mpro-x12350. Both \pearl and \boltz2 fail to place the inhibitor in the correct subpocket of the catalytic site (Figure~\ref{fig:results:rnp-qualitative}E).
Holo structures of other inhibitors in the training set occupy a similar subpocket as the \pearl and \boltz2 predicted poses. Having never seen the Mpro-x12350 subpocket during training, both models appear to have failed to generalize.

\subsubsection{Failure modes with \rmsdtwoa}
\begin{figure}[t!]
    \centering
    \includegraphics[width=.9\linewidth]
    {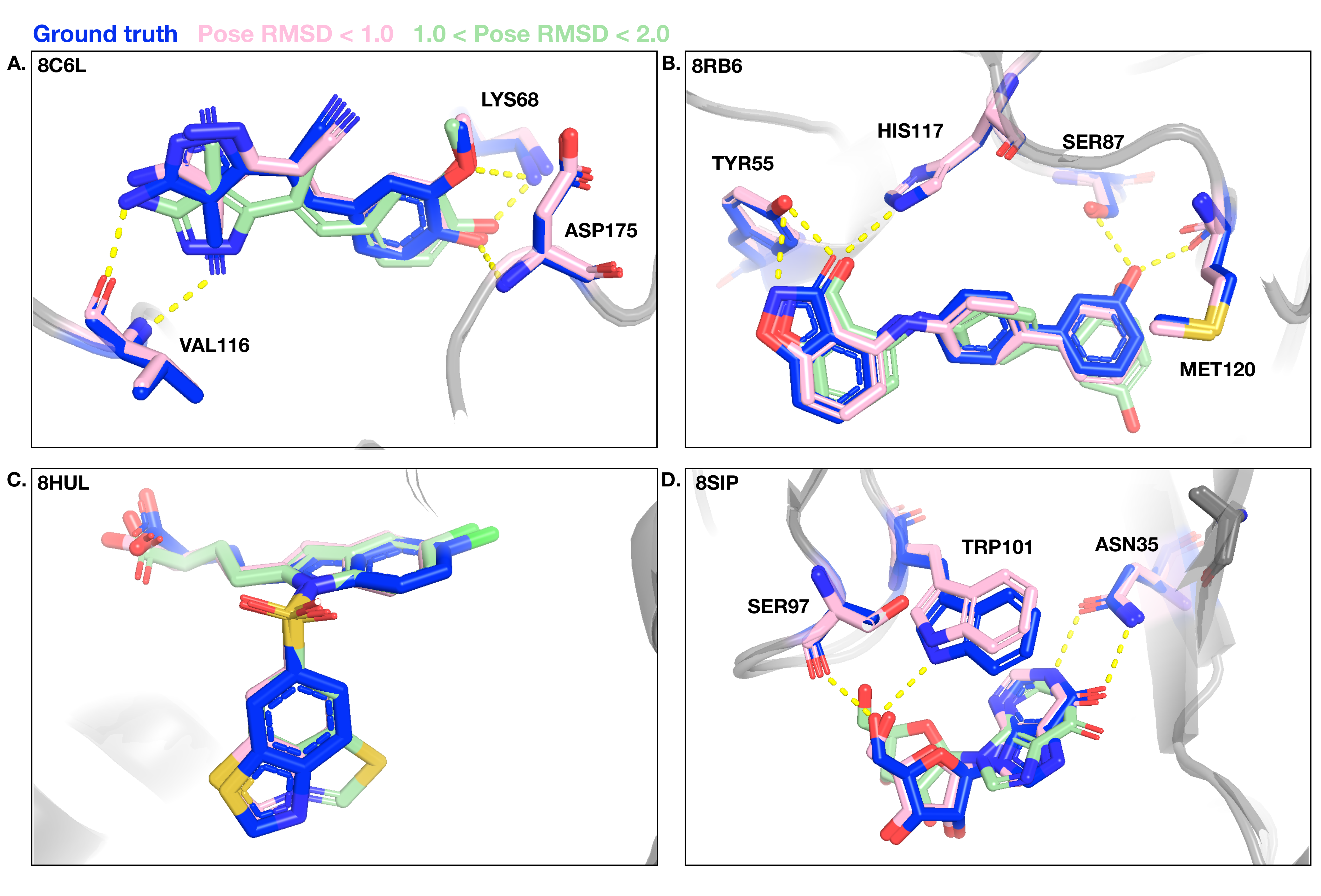}
    \caption{%
        \small
        Qualitative analysis of the importance of high-accuracy predictions. 
        The figure compares high-accuracy poses (\rmsdonea, pink) with lower-accuracy poses (\text{1\,\AA\,\raisebox{0.25ex}{\scriptsize <}} \rmsdtwoa, green). 
        The lower-accuracy poses, while under the standard \rmsdtwoa threshold, exhibit critical errors that limit their utility for drug discovery, such as ring flips (A, B, C) and slight translations (D). 
        (\textbf{A}) A 1.99~\AA pose of CK2-TN01 in complex with human protein kinase CK2 alpha displays a ring flip and misses a key interaction, which is correctly formed in the 0.42~\AA pose.
        (\textbf{B}) A 1.70~\AA pose of M689 in complex with AKR1C3 displays a problematic ring flip. Shown in comparison to the 0.36~\AA pose.
        (\textbf{C}) A 1.51~\AA pose of lanifibranor bound to human PPAR $\delta$ displays an incorrect flip of the thiazole ring. Shown in comparison to a 0.58~\AA pose.
        (\textbf{D}) A 1.54~\AA pose of inosine bound to a mouse IgG fragment displays a slight translation, missing all the key interactions that are correctly predicted in the 0.38~\AA pose.
    }
    \label{fig:results:rnp-rmsd-qualitative}
\end{figure}

\vspace{-0.75ex}
Poses below 2~\AA RMSD, the accuracy threshold used in many benchmarks, can still have significant flaws that may mislead a drug discovery scientist.
Even under this threshold, unrecoverable mistakes---such as missed key protein-ligand interactions, flipped rings leading to suboptimal interactions, or other subtle errors---can greatly reduce or even fully negate a pose's utility in drug discovery.
We highlight some of these cases, as they represent further opportunities improving cofolding models.
Figure~\ref{fig:results:rnp-rmsd-qualitative} compares \pearl's poses selected from the 20 generated samples, those that are \rmsdonea with those that are between 1--2~\AA RMSD.
Figures~\ref{fig:results:rnp-rmsd-qualitative}A, \ref{fig:results:rnp-rmsd-qualitative}B, and \ref{fig:results:rnp-rmsd-qualitative}C all show poses between 1--2~\AA RMSD that contain problematic ring flips.
Figure~\ref{fig:results:rnp-rmsd-qualitative}A shows the pyrazole ring of CK2-TN01 flipped, leading the pose to miss an important interaction with the backbone of Val116.
Similarly, we observe a pose of the inhibitor M689 bound to AKR1C3 in which a phenol flips and as a consequence misses key interactions with the side chain of Ser87 and the backbone of Met120 (Figure~\ref{fig:results:rnp-rmsd-qualitative}B).
Some ring flips lead to less obvious errors that would mislead a drug discovery scientist; we find a pose in which the thiazole ring of the PPAR $\delta$ ligand lanifibranor is flipped, but no interactions are made or broken (Figure~\ref{fig:results:rnp-rmsd-qualitative}C).
Finally, we show an example where the the pose is simply ``not quite right'' (Figure~\ref{fig:results:rnp-rmsd-qualitative}D): the inosine is very slightly translated in the pocket, leading it to miss all the key interactions with the protein.

\section{Discussion}
\label{sec:discussion}

\paragraph{Interpretation of the results}
\pearl's state-of-the-art performance arises not from a single improvement, but from the synergy between its core innovations in data, training recipe, architecture, and inference, particularly in addressing the \textit{low-data regime}, common in scientific ML. We tackle this challenge in several ways.
First, we provide strong evidence that leveraging large-scale synthetic data helps overcome PDB limitations and boosts generalization.
Crucially, we demonstrate a clear scaling relationship where model performance improves monotonically with the amount of synthetic data used in training (Figure~\ref{fig:synth-data-scaling}).
This finding validates synthetic data augmentation as a powerful strategy for building and scaling foundation models for science in data-scarce domains.
The effectiveness of this data is further enhanced by a training curriculum that progressively increases task complexity, aiding generalization.
Second, incorporating an SO(3)-equivariant diffusion module introduces a strong geometric inductive bias from the domain of geometric deep learning.
By enforcing rotational symmetry architecturally, the model achieves greater sample efficiency.
We hypothesize this contributes significantly to \pearl's improved generalization observed in generated poses.
Moreover, \pearl's substantially improved physical validity, evidenced by the minimal performance drop when PB-valid checks are applied (Section \ref{sec:results:cofolding}), suggests its architecture and training effectively enable both accuracy and physical plausibility with minimal tradeoff between the two.
Third, \pearl's multi-chain templating system aids generalization by providing richer contextual data during training and functions as a sophisticated conditioning and in-context learning framework during inference.
This allows expert users to provide structural ``prompts'' (templates) to guide generation effectively, enhancing accuracy in conditional settings (Figure~\ref{fig:results:conditional-benchmarks}).

\paragraph{Implications for drug discovery workflows}
These findings have direct implications for scientists using computational tools for structure-based drug design.
The unconditional cofolding mode suits early-stage hypothesis generation for novel targets where no known pockets or experimental structures exist.
The conditional cofolding mode is better suited for hit ID against a known pocket, or for hit-to-lead and lead optimization, where accuracy is paramount and an existing reference structure can be used to guide the model to produce poses of new or prospective analogs.

\paragraph{The need for high accuracy}
Critically, our results underscore the substantial limitations of the standard \rmsdtwoa threshold for assessing practical utility in drug discovery.
As demonstrated qualitatively (Figure~\ref{fig:results:rnp-rmsd-qualitative}) and increasingly recognized, many poses below this threshold exhibit critical flaws like incorrect ring flips or missed interactions that render them misleading for medicinal chemistry or downstream physics-based modeling. Guiding molecular design requires higher accuracy.
\pearl's significantly improved success rates at the stricter \rmsdonea threshold across all benchmarks (Figure~\ref{fig:public}, right panels) represent a key advancement.
This superior high-accuracy performance, combined with its high overall physical validity, makes \pearl's predictions substantially more reliable and genuinely useful for structure-based drug design workflows.
This practical advantage is particularly evident in demanding guided research scenarios, where \pearl's conditional cofolding mode delivers state-of-the-art accuracy even at the critical \rmsdonea threshold on both the \posebusters and \internalxtals datasets (Figure~\ref{fig:results:xrtl-rmsd-best@5-cond}).

\paragraph{Implications for AI for science} More broadly, this work serves as a case study for building generative \textit{foundation models in scientific domains}. The principles demonstrated---most notably the effectiveness of scaling with synthetic data in low-data regimes (Figure~\ref{fig:synth-data-scaling}), alongside the benefits of architectural symmetry enforcement and flexible conditioning mechanisms---are likely transferable. These techniques hold potential for tackling other complex, generative modeling problems at the intersection of ML, chemistry, physics, and biology.

\paragraph{Limitations and future work}
While \pearl significantly improves performance, particularly in novel sequence and chemical spaces (Figures \ref{fig:results:ix-rmsd-best@5-cofolding}, \ref{fig:results:xrtl-rmsd-best@5-cond}), limitations remain.
Like other models, \pearl's accuracy degrades on out-of-distribution (OOD) data, and it is susceptible to ``long-tail'' failures like mispredicting large induced-fit changes.
Memorization artifacts persist, indicating synthetic data mitigates but does not eliminate this challenge.
Furthermore, as also found by previous studies~\cite{vskrinjar2025have}, existing confidence models are often uninformative for ranking and selecting the highest-accuracy pose, as the top-ranked pose is frequently no better than a random sample (see Figure~\ref{fig:app:rnp-best-vs-maxconf} in the Appendix).
Thus, while \pearl excels at generating high-quality poses (high \bestk rates), reliably selecting the optimal pose remains a critical challenge for the field.
Future work must focus on improving OOD generalization, robustness, and developing reliable pose selection methods.
This includes exploring new training/inference protocols, better capturing protein dynamics, advanced conditioning, alternative geometric architectures, integrating physics-informed ML, and addressing computational bottlenecks.
\pearl provides a strong foundation for this next generation of generative models for science.


\section{Conclusion}
\label{sec:conclusion}

We introduced \pearl, a generative foundation model that establishes a new state-of-the-art for protein-ligand structure prediction by effectively addressing key limitations in existing cofolding systems.
We identified primary challenges in data scarcity, learning physical constraints, and limited controllability, and presented \pearl's three-pronged solution: leveraging large-scale \emph{synthetic data} within a curriculum \emph{to improve generalization and overcome PDB biases}; incorporating \emph{SO(3)-equivariant components} \emph{for enhanced sample efficiency}; and employing a flexible \emph{multi-chain templating} approach \emph{for superior controllability and conditional accuracy}.
The results demonstrate that these innovations lead to significant performance gains over existing baselines, including \alphafold3, particularly at the strictest drug discovery relevant accuracy metrics for protein-ligand complexes and small molecule tasks.

More importantly, our work goes beyond statistical benchmarks to validate the practical utility of these high-fidelity structures.
We show that poses generated by \pearl are significantly more accurate (RMSD) and physically plausible (PB-valid), rendering them tangibly more valuable for \emph{guiding medicinal chemistry decisions} and accelerating structure-based drug design workflows.
The principles established here---particularly the effectiveness of scaling with synthetic data and the benefits of equivariant design---provide a clear path forward for developing more powerful and reliable foundation models.
Continued development following these principles promises even more powerful tools for AI-driven scientific discovery.
We believe these findings are not only significant for drug discovery but also hold broader implications for generative AI at the intersection of chemistry, physics, and biology.

\section*{Acknowledgments}
\label{sec:aknowledgement}

The authors would like to thank Olivia Bass, Kathy Benemann, Ken Gong, Stacie Calad-Thomson, Kyle Tretina, and Pat Walters for their helpful feedback on the manuscript.
In addition, we thank Michael Atkin, Nicholas Barry, Rebecca Boiarsky, Elizabeth Brown, Fenghong Chen, Kush Desai, Ajinkya Deshpande, Drake Diedrich, Michael Dumont, Gianpaolo Gobbo, Alex Goldberg, Anindit Gopalakrishnan, Victoria Ingman, Rohit Kundu, Dean Latney, Shir Levkowitz, Pavel Mikhalchuk, William McCarthy, Tom Metzger, Nikhil Murthy, Parvin Parineh, Diego Puppin, Davide Sabbadin, Colter Spearsmith, Kendrick Shen, Justin Steinman, Vyom Thakkar, William Wang, Lillian Weng, Jianbo Zhao for the helpful discussion and support.

\printbibliography

\clearpage
\appendix

\section{Detailed Evaluation Methodology}
\label{app:evals}

In this section, we provide full details on our evaluation methodology, including details on the datasets and baselines (Section~\ref{app:eval:datasets}), inference modes (Section~\ref{app:eval:regimes}), and evaluation protocol and metrics (Section~\ref{app:eval:metrics}).

\subsection{Datasets and baselines}
\label{app:eval:datasets}
We evaluate \pearl against multiple cofolding baselines, including \alphafold3 \cite{abramson2024accurate} and open source descendants such as \boltz1(x)~\cite{wohlwend2024boltz1}, \boltz2~\cite{passaro2025boltz2}, \chai1~\cite{chai2024chai}, and ProteniX~\cite{bytedance2025protenix} across three benchmarks:
(i) \runsnposes~\cite{vskrinjar2025have} (which we refer to as \rnp), a recent public dataset for primary head-to-head comparisons; (ii) \posebusters~\cite{buttenschoen2024posebusters}, a standard benchmark to ensure robust performance; and (iii) \internalxtals, a curated proprietary dataset designed to test models' generalization on data representative of real-world small molecule drug discovery programs.

\paragraph{Release date cutoffs for test structures}
\pearl has been trained on crystal structures from the PDB released prior to \fmtdate{2021-09-30}.\footnote{\footnotesize Synthetic data used for training \pearl was also derived only from these publicly available structures with the same release date cutoff. No proprietary experimental data was used in training.}
\alphafold3, \boltz1(x), \chai1, and \protenix all have used release date cutoffs on or before \fmtdate{2021-09-30}.
However, \boltz2 used a later cutoff date of \fmtdate{2023-06-01}, and therefore can only be properly evaluated on proprietary structures or public structures released after that date.
For that reason, we exclude \boltz2 from the comparison on the \posebusters dataset, which contains only structures released prior to \fmtdate{2023-06-01}.
In order to compare other methods against \boltz2 on the \rnp dataset, we focus our evaluation on the subset of structures released after \fmtdate{2023-06-01}.
Analogously, we subset \posebusters to structures released on or after \fmtdate{2021-10-01} so that we can fairly compare \pearl with the other baselines.
Finally, all our models are allowed to use structures from the PDB as templates only if these structures were released prior to \fmtdate{2021-09-30} (for reference, same template cutoff used by \alphafold3 \cite{abramson2024accurate}), and more than 60 days prior to the release of the test structure.

\paragraph{Comparison with \alphafold3}
Due to the restrictive license of \alphafold3, we only evaluate \pearl against it on the \rnp benchmark, for which a third-party academic group has publicly released generated structures and computed metrics for the poses. For illustrative reasons, we include results in Figure~\ref{fig:public} and Tables~\ref{tbl:best1_metrics}--\ref{tbl:best20_metrics} for \alphafold3 for \posebusters, but using published results for poses selected for max confidence out of 25 samples (not \best5). 

\paragraph{Baselines and protocol}
\rnp provides publicly released metrics for all baselines in which we are interested; we report results for all these baselines based on this data. \rnp did not evaluate these models in the pocket conditional regime, so we do not report performance in this setting for this benchmark.
To ensure consistency of our results, for \pearl, we followed the exact inference and evaluation protocols prescribed by the authors of the benchmark \cite{vskrinjar2025have}.
For \posebusters and \internalxtals, inference has been run for \boltz1, \boltz1x, \boltz2, and \chai1 to establish baselines.

\paragraph{Processing of \rnp released metrics for consistency}
Even though \rnp authors aimed to generate 25 poses per test structure (5 seeds $\times$ 5 samples) for each evaluated cofolding models, they reported that different models failed on some of the structures.
In our analysis, we noticed that some models not only failed to produce poses entirely for some of the structures, but the released metrics had an inconsistent number of poses per structure.%
\footnote{\footnotesize We found cases where \runsnposes released data for each model that contained some structures with fewer than 25 poses generated, ranging from as few as 1--5 poses to often between 20 and 24 poses per structure.}
To ensure consistency of our analysis, in addition to imposing \fmtdate{2023-06-01+} release date cutoff, we processed \rnp data in the following manner: (i) we only considered a subset of structures for which all methods had least 20 poses, (ii) we selected the first 20 generated poses (4 seeds $\times$ 5 samples) for each structure for each baseline method.
This enabled a fully consistent comparison with \pearl for which we generated 20 poses per structure.
This resulting \rnp test set contained 702 structures.

\begin{figure}[t]
    \centering
    \begin{minipage}[b]{0.48\linewidth}
        \small
        \begin{tabular}{@{}l|r|r@{}}
            \toprule
            \textbf{Dataset} & \textbf{Structure release date} & \textbf{\# structures} \\ \midrule
            \runsnposes & \fmtdate{2023-06-01+} & 702 \\
            \posebusters & \fmtdate{2021-10-01+} & 297 \\
            \internalxtals & proprietary & 111 \\
            \bottomrule
        \end{tabular}
        \vspace{5.0ex}
    \end{minipage}
    \hfill
    \begin{minipage}[b]{0.48\linewidth}
        \includegraphics[width=\linewidth]{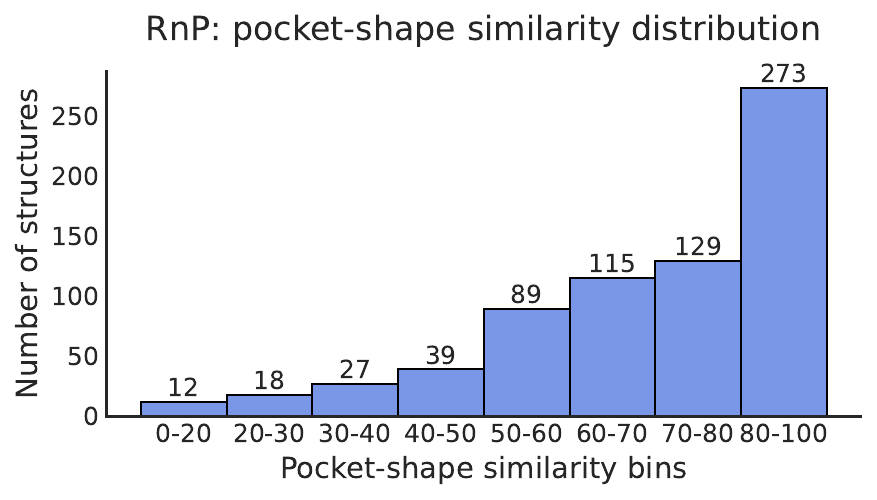}
    \end{minipage}
    \caption{(Left) statistics of the \rnp, \posebusters, and \internalxtals datasets used for analysis and benchmarking. (Right) Histogram of the number of structures in each pocket-shape similarity bin of the \rnp dataset.}
    \label{fig:dataset-statistics}
\end{figure}

\paragraph{Size and composition of the datasets}
The keys statistics of the final benchmarking sets we used in our analysis (after filtering and processing) are summarized in Figure~\ref{fig:dataset-statistics} (left).
For each single structure in each of the datasets, we ensured that each cofolding model was able to produce 20 poses for which metrics were computed (for \rnp, we used metrics released by the authors).
Figure~\ref{fig:dataset-statistics} (right) shows the distribution over pocket-shape similarity bins of the final \rnp dataset used in our analysis.

\subsection{Unconditional and conditional evaluation regimes}
\label{app:eval:regimes}

To assess flexibility and performance in different scenarios, we evaluate \pearl in two distinct experimental setups: (i) the unconditional cofolding and (ii) the pocket-conditional cofolding modes.

\textbf{The unconditional cofolding mode} tests the model's ability to generate a complex structure starting from only the protein's amino acid sequence and the ligand's 2D structure, as well as MSA information.
This evaluation assumes that no prior structural information about the target is available and the binding pocket is unknown.
Consequently, any template search must rely on sequence and ligand similarity alone.
As the publicly available version of \alphafold3 operates in this mode, our primary comparison against it is in this regime.

\textbf{The conditional cofolding mode} is designed to emulate a common scenario in drug discovery, where a reference structure (from public sources or proprietary crystal data) for the target protein is available and a specific binding pocket is hypothesized.
This setup evaluates \pearl's controllability and its performance in practical, structure-guided scenarios where a scientist provides a structural prior to guide the generation process.
Our evaluation in this regime includes \pearl and the open source baselines that support this type of conditioning: \boltz1(x), \boltz2, and \chai1.
More concretely, each of the evaluated models is given access to some of the residues that constitute the binding pocket, which are then provided to the models at inference time.
For the purposes of automated benchmarking, the pocket residues must be carefully selected to avoid unfairly leaking information.
To do so, we select at most two residues.
To select the first, the distances to any residue within 6~\AA is calculated for each ligand atom, and the residue with the smallest median distance is selected.
The same procedure is used to find the next best residue that is separated in protein sequence from the first by at least eight residues.
The goal is to approximately select two residues from opposing sides of the pocket with the target benchmark ligand reasonably centered between them.
For each test complex, the identities of the same $\leq2$ pocket residues were provided to all evaluated models.

\subsection{Evaluation protocol and metrics}
\label{app:eval:metrics}
Our protocol is designed to provide a multifaceted view of model performance.

\textbf{Pose Accuracy:}
Our primary metric for structural accuracy is ligand RMSD (also known as BiSyRMSD, or Binding-Site Superposed, Symmetry-Corrected Pose RMSD) \cite{robin2023assessment}.
We report the industry standard success rates at specific quality thresholds (\eg, \rmsdtwoa and \rmsdonea).
Our main results are focused on ligand \rmsdtwoa and \rmsdonea success rates.
We report additional metrics in Appendix~\ref{app:results}.

\textbf{Pose Quality:}
To ensure poses are physically plausible, we use PoseBusters' plausibility checks in the ``redock'' mode using the PoseBusters package version \texttt{0.2.9} \cite{buttenschoen2024posebusters}. This includes the following checks. A pose is considered valid (PB-valid) only if it passes all checks.
\begin{multicols}{2} 
\small
\begin{itemize}
    \item \texttt{mol\_pred\_loaded}
    \item \texttt{mol\_true\_loaded}
    \item \texttt{mol\_cond\_loaded}
    \item \texttt{sanitization}
    \item \texttt{all\_atoms\_connected}
    \item \texttt{molecular\_formula}
    \item \texttt{molecular\_bonds}
    \item \texttt{double\_bond\_stereochemistry}
    \item \texttt{tetrahedral\_chirality}
    \item \texttt{bond\_lengths}
    \item \texttt{bond\_angles}
    \item \texttt{internal\_steric\_clash}
    \item \texttt{aromatic\_ring\_flatness}
    \item \texttt{double\_bond\_flatness}
    \item \texttt{internal\_energy}
    \item \texttt{protein-ligand\_maximum\_distance}
    \item \texttt{minimum\_distance\_to\_protein}
    \item \texttt{minimum\_distance\_to\_organic\_cofactors}
    \item \texttt{minimum\_distance\_to\_inorganic\_cofactors}
    \item \texttt{minimum\_distance\_to\_waters}
    \item \texttt{volume\_overlap\_with\_protein}
    \item \texttt{volume\_overlap\_with\_organic\_cofactors}
    \item \texttt{volume\_overlap\_with\_inorganic\_cofactors}
    \item \texttt{volume\_overlap\_with\_waters}
\end{itemize}
\end{multicols}

\textbf{Uncertainty estimates via bootstrapping:}
We bootstrap \cite{efron1994introduction} with 1000 iterations to estimate the mean and standard error over the complexes in each dataset for all metrics computed in our study.
Unless specified otherwise, all error bars on the reported charts and tables correspond to standard error of the mean.

\begin{figure}
    \centering
    \includegraphics[width=1.0\linewidth]{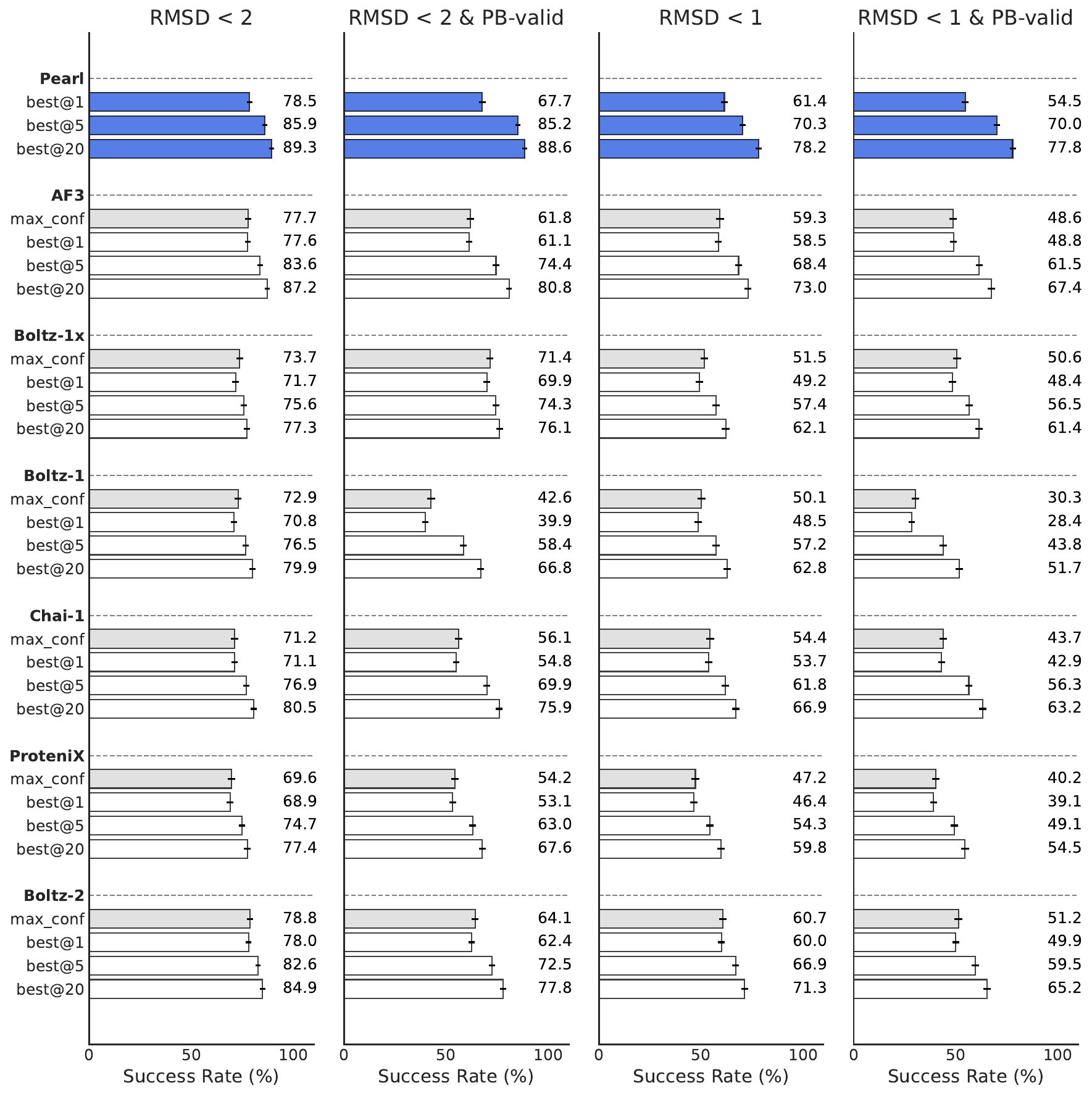}
    \caption{%
        \small
        Comparison of the success rates for the \best1, \best5, and \best{20} poses versus poses selected with a confidence model (out of 20 sampled) for each of the models on the \runsnposes benchmark.
        The differences between max confidence and randomly selected (\best1) poses are not statistically significant for any of the models.
        In many cases, \pearl's \best1 pose (\ie, a single random pose) outperforms max confidence poses produced by other models.
    }
    \label{fig:app:rnp-best-vs-maxconf}
\end{figure}

\subsection{The \bestk evaluation protocol vs pose selection with a confidence model}
\label{app:experimental-setup:bestk}

Given the input information about the system of interest, cofolding models can generate multiple pose candidates.
Some of the generated poses can be more accurate and higher quality than others.
In cofolding, some of the previous benchmarking efforts computed confidence scores and reported metrics on the highest confidence pose \cite{vskrinjar2025have}.
However, we see \emph{pose generation} and \emph{pose selection} as two different problems, which should not be conflated.
Instead of evaluating only the model's top-ranked pose by a confidence score, we adopt the common practice from generative modeling literature and compute metrics for the \bestk poses, where \best5 is one of the standards in the literature \cite{chen2021evaluating}.
In essence, \bestk means that $k$ random samples are generated from the model and the best one is selected according to the metric of interest.%
\footnote{\footnotesize In practice, we pre-generate \(n = 20\) poses for each test complex and use an unbiased estimator for \bestk as proposed in \cite{chen2021evaluating}.}

\bestk metrics have a couple of important properties:
\begin{enumerate}
    \item \textbf{\best1 behaves similar to \emph{precision}.}
    If the metric is a binary indicator of success (\eg, \rmsdtwoa), \best1 returns the expected probability of success of a random sample generated by the model.
    \item \textbf{\bestk for larger $k$ values behave similar to \texttt{recall@k}.}
    Again, for a binary indicator of success, \bestk returns the expected probability of generating a successful sample after $k$ attempts.
\end{enumerate}
Which $k$ is more important in practice depends on the setting of interest.
For example, if a user wanted to guarantee that every pose generated by the model has high probability of success, it's important to optimize for \texttt{best@1}.
On the other hand, if a user has access to an appropriate scoring method, and computational cost of inference is not rate-limiting, one should optimize for the probability to ``recall'' a high quality pose given enough attempts, and use \bestk with larger $k$.
Our main results focus on \best5 evaluation, which is a standard metric in the broader generative modeling literature, and it strikes a good balance between \best1 and \best{20}.

\paragraph{Comparison of \bestk with pose selection using confidence model}
To understand the differences between \best1, \best5, \best{20}, and top-ranked poses selected by confidence (out of 20 generated), we computed ligand RMSD metrics for each of these pose aggregation methods on the \rnp dataset (which includes confidence scores for each pose for each baseline method).
The results are presented in Figure~\ref{fig:app:rnp-best-vs-maxconf}.
Surprisingly, we found that confidence scores are not particularly informative for ranking poses of protein-ligand interfaces: the results show that the top-ranked pose by confidence score is no better than a randomly selected pose on (\best1); the difference between max confidence and \best1 is not statistically significant for any of the models.
Our finding is consistent with the observations made in \cite[][Section 2.5]{vskrinjar2025have}.
We also note that random poses (\best1) generated by \pearl in many cases are superior compared to max confidence poses produced by other models.

\section{Extended results}
\label{app:results}

Below, we present additional results for \pearl and baseline cofolding models on the \rnp, \posebusters, and \internalxtals benchmarks and provide further analysis and discussion.

\subsection{Comprehensive metrics}
\label{app:metrics}

\newcommand{\pmTiny}{\raisebox{0.3ex}{\tiny $\pm$}\xspace}

\begin{table}[htbp]
\centering
\small
\begin{threeparttable}
\caption{Model performance at \textbf{\texttt{best@1}} aggregation.}
\label{tbl:best1_metrics}
\renewcommand{\arraystretch}{1.1}
\begin{tabular}{l l l | r r r r r}
\toprule
Mode & Dataset & Method & \rmsdtwo & \makecell[r]{\rmsdtwo\\\& PB-valid} & \rmsdone & \makecell[r]{\rmsdone\\\& PB-valid} & lDDT-PLI \\
\midrule
\multirow{17}{*}{\rotatebox[origin=c]{90}{\textit{Unconditional}}}
 & \multirow{7}{*}{\runsnposes} & \pearl & \textbf{79.0 \pmTiny 1.4} & 67.9 \pmTiny 1.6 & \textbf{61.8 \pmTiny 1.7} & \textbf{55.0 \pmTiny 1.7} & \textbf{81.9 \pmTiny 0.9} \\
 &  & \af{3} & 77.6 \pmTiny 1.4 & 61.1 \pmTiny 1.5 & 58.5 \pmTiny 1.7 & 48.8 \pmTiny 1.6 & 80.9 \pmTiny 0.9 \\
 &  & \boltz{1x} & 71.7 \pmTiny 1.6 & \textbf{69.9 \pmTiny 1.7} & 49.2 \pmTiny 1.8 & 48.4 \pmTiny 1.8 & 76.9 \pmTiny 1.0 \\
 &  & \boltz{1} & 70.8 \pmTiny 1.6 & 39.9 \pmTiny 1.5 & 48.5 \pmTiny 1.8 & 28.4 \pmTiny 1.4 & 76.5 \pmTiny 0.9 \\
 &  & \chai{1} & 71.1 \pmTiny 1.6 & 54.8 \pmTiny 1.6 & 53.7 \pmTiny 1.7 & 42.9 \pmTiny 1.6 & 76.5 \pmTiny 1.0 \\
 &  & \protenix & 68.9 \pmTiny 1.6 & 53.1 \pmTiny 1.6 & 46.4 \pmTiny 1.8 & 39.1 \pmTiny 1.6 & 76.0 \pmTiny 0.9 \\
\cmidrule(lr){3-8}
 &  & \boltz{2}\textsuperscript{\dag} (2023-06) & 78.0 \pmTiny 1.4 & 62.4 \pmTiny 1.6 & 60.0 \pmTiny 1.7 & 49.9 \pmTiny 1.7 & 80.6 \pmTiny 0.9 \\
\cmidrule(lr){2-8}
 & \multirow{5}{*}{\posebusters} & \pearl & \textbf{79.3 \pmTiny 2.3} & \textbf{65.4 \pmTiny 2.7} & \textbf{62.4 \pmTiny 2.7} & \textbf{54.7 \pmTiny 2.7} & \textbf{82.1 \pmTiny 1.4} \\
 &  & \af{3}\textsuperscript{*} & 76.5 \pmTiny 2.5 & 60.4 \pmTiny 2.8 & 62.1 \pmTiny 2.8 & 52.7 \pmTiny 2.9 & --- \\
 &  & \boltz{1x} & 66.5 \pmTiny 2.5 & 64.1 \pmTiny 2.5 & 41.4 \pmTiny 2.5 & 40.2 \pmTiny 2.4 & 71.9 \pmTiny 1.5 \\
 &  & \boltz{1} & 66.6 \pmTiny 2.6 & 41.6 \pmTiny 2.5 & 43.8 \pmTiny 2.7 & 30.3 \pmTiny 2.4 & 72.9 \pmTiny 1.5 \\
 &  & \chai{1} & 69.6 \pmTiny 2.4 & 53.7 \pmTiny 2.5 & 51.5 \pmTiny 2.5 & 42.7 \pmTiny 2.4 & 74.7 \pmTiny 1.6 \\
\cmidrule(lr){2-8}
 & \multirow{5}{*}{\internalxtals} & \pearl & \textbf{63.2 \pmTiny 4.4} & \textbf{40.2 \pmTiny 4.1} & \textbf{29.5 \pmTiny 3.5} & \textbf{22.0 \pmTiny 3.3} & \textbf{63.8 \pmTiny 3.7} \\
 &  & \boltz{1x} & 33.1 \pmTiny 4.0 & 28.6 \pmTiny 3.7 & 3.8 \pmTiny 1.2 & 3.6 \pmTiny 1.2 & 53.8 \pmTiny 3.2 \\
 &  & \boltz{1} & 32.8 \pmTiny 4.1 & 7.5 \pmTiny 1.9 & 4.8 \pmTiny 1.5 & 1.4 \pmTiny 0.6 & 54.4 \pmTiny 3.3 \\
 &  & \chai{1} & 15.5 \pmTiny 3.3 & 5.5 \pmTiny 2.0 & 5.4 \pmTiny 1.7 & 1.6 \pmTiny 0.8 & 38.7 \pmTiny 2.8 \\
\cmidrule(lr){3-8}
 &  & \boltz{2}\textsuperscript{\dag} (2023-06) & 59.3 \pmTiny 4.3 & 29.2 \pmTiny 3.9 & 12.9 \pmTiny 2.6 & 8.6 \pmTiny 2.2 & 62.3 \pmTiny 3.6 \\
\midrule
\multirow{9}{*}{\rotatebox[origin=c]{90}{\textit{Conditional}}}
 & \multirow{4}{*}{\posebusters} & \pearl & \textbf{80.6 \pmTiny 2.2} & 66.8 \pmTiny 2.6 & \textbf{62.9 \pmTiny 2.6} & \textbf{55.2 \pmTiny 2.7} & \textbf{84.1 \pmTiny 1.1} \\
 &  & \boltz{1x} & 69.4 \pmTiny 2.5 & \textbf{67.0 \pmTiny 2.5} & 45.1 \pmTiny 2.5 & 43.9 \pmTiny 2.4 & 75.4 \pmTiny 1.2 \\
 &  & \boltz{1} & 69.6 \pmTiny 2.6 & 43.9 \pmTiny 2.6 & 47.6 \pmTiny 2.7 & 32.5 \pmTiny 2.4 & 76.4 \pmTiny 1.2 \\
 &  & \chai{1} & 72.8 \pmTiny 2.3 & 56.1 \pmTiny 2.5 & 55.1 \pmTiny 2.5 & 45.6 \pmTiny 2.4 & 77.9 \pmTiny 1.3 \\
\cmidrule(lr){2-8}
 & \multirow{5}{*}{\internalxtals} & \pearl & \textbf{68.0 \pmTiny 4.1} & \textbf{42.6 \pmTiny 4.0} & \textbf{30.7 \pmTiny 3.6} & \textbf{23.6 \pmTiny 3.4} & \textbf{75.1 \pmTiny 2.3} \\
 &  & \boltz{1x} & 42.9 \pmTiny 4.0 & 37.8 \pmTiny 3.8 & 5.2 \pmTiny 1.6 & 5.0 \pmTiny 1.6 & 62.1 \pmTiny 3.1 \\
 &  & \boltz{1} & 43.5 \pmTiny 4.2 & 10.0 \pmTiny 2.0 & 6.6 \pmTiny 1.8 & 1.8 \pmTiny 0.7 & 63.8 \pmTiny 3.0 \\
 &  & \chai{1} & 25.0 \pmTiny 3.7 & 9.2 \pmTiny 2.2 & 6.1 \pmTiny 1.7 & 1.9 \pmTiny 0.8 & 54.3 \pmTiny 2.5 \\
\cmidrule(lr){3-8}
 &  & \boltz{2}\textsuperscript{\dag} (2023-06) & 60.4 \pmTiny 4.3 & 30.9 \pmTiny 3.9 & 14.2 \pmTiny 2.7 & 9.8 \pmTiny 2.3 & 69.7 \pmTiny 2.6 \\
\bottomrule
\end{tabular}
\begin{tablenotes}
\item[*] \alphafold3 results for \posebusters use officially released metrics for poses selected for max confidence out of 25 samples (not \best{1}).
\item[\dag] Since it is trained with data up to \fmtdate{2023-06-01}, \boltz2 is not directly comparable to the other evaluated models, which in contrast all use earlier training cutoffs of $\leq$ \fmtdate{2021-09-30}.
\end{tablenotes}
\end{threeparttable}
\end{table}

\begin{table}[htbp]
\centering
\small
\begin{threeparttable}
\caption{Model performance at \textbf{\texttt{best@5}} aggregation.}
\label{tbl:best5_metrics}
\renewcommand{\arraystretch}{1.1}
\begin{tabular}{l l l | r r r r r}
\toprule
Mode & Dataset & Method & \rmsdtwo & \makecell[r]{\rmsdtwo\\\& PB-valid} & \rmsdone & \makecell[r]{\rmsdone\\\& PB-valid} & lDDT-PLI \\
\midrule
\multirow{17}{*}{\rotatebox[origin=c]{90}{\textit{Unconditional}}}
 & \multirow{7}{*}{\runsnposes} & \pearl & \textbf{85.9 \pmTiny 1.2} & \textbf{85.2 \pmTiny 1.2} & \textbf{70.3 \pmTiny 1.5} & \textbf{70.0 \pmTiny 1.6} & \textbf{84.6 \pmTiny 0.7} \\
 &  & \af{3} & 83.6 \pmTiny 1.3 & 74.4 \pmTiny 1.5 & 68.4 \pmTiny 1.7 & 61.5 \pmTiny 1.7 & 84.4 \pmTiny 0.8 \\
 &  & \boltz{1x} & 75.6 \pmTiny 1.6 & 74.3 \pmTiny 1.6 & 57.4 \pmTiny 1.8 & 56.5 \pmTiny 1.8 & 79.5 \pmTiny 0.9 \\
 &  & \boltz{1} & 76.5 \pmTiny 1.5 & 58.4 \pmTiny 1.7 & 57.2 \pmTiny 1.8 & 43.8 \pmTiny 1.7 & 79.9 \pmTiny 0.9 \\
 &  & \chai{1} & 76.9 \pmTiny 1.5 & 69.9 \pmTiny 1.6 & 61.8 \pmTiny 1.7 & 56.3 \pmTiny 1.8 & 80.1 \pmTiny 1.0 \\
 &  & \protenix & 74.7 \pmTiny 1.6 & 63.0 \pmTiny 1.7 & 54.3 \pmTiny 1.8 & 49.1 \pmTiny 1.8 & 79.5 \pmTiny 0.9 \\
\cmidrule(lr){3-8}
 &  & \boltz{2}\textsuperscript{\dag}~(2023-06) & 82.6 \pmTiny 1.3 & 72.5 \pmTiny 1.5 & 66.9 \pmTiny 1.7 & 59.5 \pmTiny 1.7 & 83.4 \pmTiny 0.8 \\
\cmidrule(lr){2-8}
 & \multirow{5}{*}{\posebusters} & \pearl & \textbf{85.1 \pmTiny 2.0} & \textbf{84.7 \pmTiny 2.0} & \textbf{72.7 \pmTiny 2.4} & \textbf{72.4 \pmTiny 2.4} & \textbf{84.9 \pmTiny 1.1} \\
 &  & \af{3}\textsuperscript{*} & 76.5 \pmTiny 2.5 & 60.4 \pmTiny 2.8 & 62.1 \pmTiny 2.8 & 52.7 \pmTiny 2.9 & --- \\
 &  & \boltz{1x} & 74.8 \pmTiny 2.4 & 74.2 \pmTiny 2.4 & 55.6 \pmTiny 2.7 & 55.2 \pmTiny 2.7 & 77.7 \pmTiny 1.3 \\
 &  & \boltz{1} & 73.2 \pmTiny 2.5 & 54.5 \pmTiny 2.8 & 54.0 \pmTiny 2.8 & 41.9 \pmTiny 2.7 & 77.5 \pmTiny 1.4 \\
 &  & \chai{1} & 77.8 \pmTiny 2.3 & 68.7 \pmTiny 2.5 & 63.6 \pmTiny 2.5 & 58.0 \pmTiny 2.6 & 79.8 \pmTiny 1.4 \\
\cmidrule(lr){2-8}
 & \multirow{5}{*}{\internalxtals} & \pearl & \textbf{66.1 \pmTiny 4.4} & \textbf{62.0 \pmTiny 4.4} & \textbf{37.2 \pmTiny 3.7} & \textbf{36.0 \pmTiny 3.7} & \textbf{64.3 \pmTiny 3.5} \\
 &  & \boltz{1x} & 39.4 \pmTiny 4.4 & 36.6 \pmTiny 4.3 & 9.9 \pmTiny 2.3 & 9.3 \pmTiny 2.3 & 55.6 \pmTiny 3.3 \\
 &  & \boltz{1} & 37.6 \pmTiny 4.3 & 14.5 \pmTiny 3.0 & 9.9 \pmTiny 2.5 & 4.2 \pmTiny 1.5 & 55.9 \pmTiny 3.3 \\
 &  & \chai{1} & 17.1 \pmTiny 3.5 & 6.7 \pmTiny 2.2 & 9.5 \pmTiny 2.6 & 3.2 \pmTiny 1.6 & 40.9 \pmTiny 2.8 \\
\cmidrule(lr){3-8}
 &  & \boltz{2}\textsuperscript{\dag} (2023-06) & 64.9 \pmTiny 4.4 & 38.0 \pmTiny 4.3 & 22.3 \pmTiny 3.4 & 14.6 \pmTiny 2.9 & 63.8 \pmTiny 3.6 \\
\midrule
\multirow{9}{*}{\rotatebox[origin=c]{90}{\textit{Conditional}}}
 & \multirow{4}{*}{\posebusters} & \pearl & \textbf{86.7 \pmTiny 1.8} & \textbf{86.7 \pmTiny 1.8} & \textbf{72.6 \pmTiny 2.4} & \textbf{72.2 \pmTiny 2.4} & \textbf{86.2 \pmTiny 0.9} \\
 &  & \boltz{1x} & 76.8 \pmTiny 2.3 & 76.3 \pmTiny 2.3 & 59.5 \pmTiny 2.7 & 59.4 \pmTiny 2.7 & 80.5 \pmTiny 1.1 \\
 &  & \boltz{1} & 75.6 \pmTiny 2.4 & 57.1 \pmTiny 2.8 & 57.8 \pmTiny 2.8 & 44.7 \pmTiny 2.7 & 80.5 \pmTiny 1.1 \\
 &  & \chai{1} & 81.4 \pmTiny 2.1 & 70.7 \pmTiny 2.4 & 67.7 \pmTiny 2.5 & 60.6 \pmTiny 2.6 & 82.9 \pmTiny 1.2 \\
\cmidrule(lr){2-8}
 & \multirow{5}{*}{\internalxtals} & \pearl & \textbf{81.4 \pmTiny 3.3} & \textbf{73.9 \pmTiny 3.8} & \textbf{41.1 \pmTiny 3.7} & \textbf{39.8 \pmTiny 3.8} & \textbf{81.4 \pmTiny 1.4} \\
 &  & \boltz{1x} & 54.9 \pmTiny 4.4 & 51.5 \pmTiny 4.4 & 11.2 \pmTiny 2.5 & 10.6 \pmTiny 2.5 & 66.2 \pmTiny 3.0 \\
 &  & \boltz{1} & 52.3 \pmTiny 4.3 & 21.9 \pmTiny 3.5 & 13.0 \pmTiny 2.8 & 5.4 \pmTiny 1.7 & 67.7 \pmTiny 2.9 \\
 &  & \chai{1} & 32.3 \pmTiny 4.1 & 17.2 \pmTiny 3.1 & 11.9 \pmTiny 2.9 & 4.5 \pmTiny 1.8 & 60.5 \pmTiny 2.4 \\
\cmidrule(lr){3-8}
 &  & \boltz{2}\textsuperscript{\dag} (2023-06) & 66.8 \pmTiny 4.3 & 41.2 \pmTiny 4.4 & 25.0 \pmTiny 3.6 & 18.1 \pmTiny 3.2 & 74.0 \pmTiny 2.3 \\
\bottomrule
\end{tabular}
\begin{tablenotes}
\item[*] \alphafold3 results for \posebusters use officially released metrics for poses selected for max confidence out of 25 samples (not \best{5}).
\item[\dag] Since it is trained with data up to \fmtdate{2023-06-01}, \boltz2 is not directly comparable to the other evaluated models, which in contrast all use earlier training cutoffs of $\leq$ \fmtdate{2021-09-30}.
\end{tablenotes}
\end{threeparttable}
\end{table}

\begin{table}[htbp]
\centering
\small
\begin{threeparttable}
\caption{Model performance at \textbf{\texttt{best@20}} aggregation.}
\label{tbl:best20_metrics}
\renewcommand{\arraystretch}{1.1}
\begin{tabular}{l l l | r r r r r}
\toprule
Mode & Dataset & Method & \rmsdtwo & \makecell[r]{\rmsdtwo\\\& PB-valid} & \rmsdone & \makecell[r]{\rmsdone\\\& PB-valid} & lDDT-PLI \\
\midrule
\multirow{17}{*}{\rotatebox[origin=c]{90}{\textit{Unconditional}}}
 & \multirow{7}{*}{\runsnposes} & \pearl & \textbf{89.3 \pmTiny 1.2} & \textbf{88.6 \pmTiny 1.2} & \textbf{78.2 \pmTiny 1.5} & \textbf{77.8 \pmTiny 1.5} & \textbf{87.4 \pmTiny 0.6} \\
 &  & \af{3} & 87.2 \pmTiny 1.2 & 80.8 \pmTiny 1.5 & 73.0 \pmTiny 1.7 & 67.4 \pmTiny 1.8 & 86.3 \pmTiny 0.8 \\
 &  & \boltz{1x} & 77.3 \pmTiny 1.6 & 76.1 \pmTiny 1.6 & 62.1 \pmTiny 1.9 & 61.4 \pmTiny 1.9 & 80.8 \pmTiny 0.9 \\
 &  & \boltz{1} & 79.9 \pmTiny 1.5 & 66.8 \pmTiny 1.8 & 62.8 \pmTiny 1.9 & 51.7 \pmTiny 1.9 & 81.8 \pmTiny 0.8 \\
 &  & \chai{1} & 80.5 \pmTiny 1.5 & 75.9 \pmTiny 1.6 & 66.9 \pmTiny 1.8 & 63.2 \pmTiny 1.9 & 82.2 \pmTiny 0.9 \\
 &  & \protenix & 77.4 \pmTiny 1.6 & 67.6 \pmTiny 1.7 & 59.8 \pmTiny 1.9 & 54.5 \pmTiny 1.9 & 81.6 \pmTiny 0.9 \\
\cmidrule(lr){3-8}
 &  & \boltz{2}\textsuperscript{\dag} (2023-06) & 84.9 \pmTiny 1.3 & 77.8 \pmTiny 1.5 & 71.3 \pmTiny 1.7 & 65.2 \pmTiny 1.8 & 85.2 \pmTiny 0.8 \\
\cmidrule(lr){2-8}
 & \multirow{5}{*}{\posebusters} & \pearl & \textbf{88.3 \pmTiny 1.9} & \textbf{88.3 \pmTiny 1.9} & \textbf{79.8 \pmTiny 2.4} & \textbf{79.1 \pmTiny 2.4} & \textbf{87.9 \pmTiny 1.0} \\
 &  & \af{3}\textsuperscript{*} & 76.5 \pmTiny 2.5 & 60.4 \pmTiny 2.8 & 62.1 \pmTiny 2.8 & 52.7 \pmTiny 2.9 & --- \\
 &  & \boltz{1x} & 80.1 \pmTiny 2.3 & 80.1 \pmTiny 2.3 & 65.8 \pmTiny 2.8 & 65.8 \pmTiny 2.8 & 81.1 \pmTiny 1.2 \\
 &  & \boltz{1} & 78.0 \pmTiny 2.5 & 62.0 \pmTiny 2.9 & 61.7 \pmTiny 2.9 & 50.1 \pmTiny 3.0 & 80.5 \pmTiny 1.3 \\
 &  & \chai{1} & 81.1 \pmTiny 2.3 & 76.0 \pmTiny 2.5 & 69.9 \pmTiny 2.7 & 65.8 \pmTiny 2.8 & 82.1 \pmTiny 1.4 \\
\cmidrule(lr){2-8}
 & \multirow{5}{*}{\internalxtals} & \pearl & \textbf{67.7 \pmTiny 4.4} & \textbf{64.1 \pmTiny 4.5} & \textbf{52.4 \pmTiny 4.7} & \textbf{50.6 \pmTiny 4.7} & \textbf{66.9 \pmTiny 3.5} \\
 &  & \boltz{1x} & 43.5 \pmTiny 4.6 & 40.8 \pmTiny 4.6 & 19.1 \pmTiny 3.8 & 17.3 \pmTiny 3.6 & 56.8 \pmTiny 3.4 \\
 &  & \boltz{1} & 43.5 \pmTiny 4.6 & 21.0 \pmTiny 3.8 & 14.6 \pmTiny 3.4 & 8.30 \pmTiny 2.6 & 57.0 \pmTiny 3.4 \\
 &  & \chai{1} & 20.0 \pmTiny 3.8 & 9.10 \pmTiny 2.8 & 12.8 \pmTiny 3.2 & 4.60 \pmTiny 2.0 & 42.8 \pmTiny 2.9 \\
\cmidrule(lr){3-8}
 &  & \boltz{2}\textsuperscript{\dag} (2023-06) & 65.9 \pmTiny 4.4 & 46.0 \pmTiny 4.7 & 34.3 \pmTiny 4.5 & 23.4 \pmTiny 4.0 & 64.6 \pmTiny 3.7 \\
\midrule
\multirow{9}{*}{\rotatebox[origin=c]{90}{\textit{Conditional}}}
 & \multirow{4}{*}{\posebusters} & \pearl & \textbf{91.1 \pmTiny 1.6} & \textbf{91.1 \pmTiny 1.6} & \textbf{79.8 \pmTiny 2.4} & \textbf{79.4 \pmTiny 2.4} & \textbf{89.1 \pmTiny 0.8} \\
 &  & \boltz{1x} & 83.5 \pmTiny 2.2 & 83.5 \pmTiny 2.2 & 67.2 \pmTiny 2.8 & 67.2 \pmTiny 2.8 & 83.6 \pmTiny 1.0 \\
 &  & \boltz{1} & 80.8 \pmTiny 2.3 & 64.1 \pmTiny 2.8 & 63.4 \pmTiny 2.9 & 51.8 \pmTiny 2.9 & 82.9 \pmTiny 1.1 \\
 &  & \chai{1} & 84.5 \pmTiny 2.1 & 76.8 \pmTiny 2.4 & 73.9 \pmTiny 2.5 & 65.8 \pmTiny 2.7 & 85.1 \pmTiny 1.1 \\
\cmidrule(lr){2-8}
 & \multirow{5}{*}{\internalxtals} & \pearl & \textbf{88.3 \pmTiny 3.0} & \textbf{79.2 \pmTiny 3.9} & \textbf{60.4 \pmTiny 4.7} & \textbf{57.7 \pmTiny 4.8} & \textbf{85.9 \pmTiny 1.1} \\
 &  & \boltz{1x} & 60.4 \pmTiny 4.6 & 57.7 \pmTiny 4.7 & 19.9 \pmTiny 3.8 & 18.0 \pmTiny 3.7 & 68.8 \pmTiny 2.9 \\
 &  & \boltz{1} & 59.6 \pmTiny 4.6 & 29.9 \pmTiny 4.3 & 20.0 \pmTiny 3.9 & 10.1 \pmTiny 2.9 & 70.4 \pmTiny 2.7 \\
 &  & \chai{1} & 39.1 \pmTiny 4.6 & 25.4 \pmTiny 4.1 & 15.5 \pmTiny 3.5 & 5.6 \pmTiny 2.2 & 62.7 \pmTiny 2.4 \\
\cmidrule(lr){3-8}
 &  & \boltz{2}\textsuperscript{\dag} (2023-06) & 69.5 \pmTiny 4.3 & 48.8 \pmTiny 4.8 & 35.3 \pmTiny 4.5 & 26.4 \pmTiny 4.2 & 76.8 \pmTiny 2.1 \\
\bottomrule
\end{tabular}
\begin{tablenotes}
\item[*] \alphafold3 results for \posebusters use officially released metrics for poses selected for max confidence out of 25 samples (not \best{20}).
\item[\dag] Since it is trained with data up to \fmtdate{2023-06-01}, \boltz2 is not directly comparable to the other evaluated models, which in contrast all use earlier training cutoffs of $\leq$ \fmtdate{2021-09-30}.
\end{tablenotes}
\end{threeparttable}
\end{table}

Tables \ref{tbl:best1_metrics}, \ref{tbl:best5_metrics}, and \ref{tbl:best20_metrics} present the extended set of metrics for \best{1}, \best{5}, and \best{20} pose aggregation, respectively.
For each pose aggregation level, we provide detailed results across inference modes (unconditional and pocket-conditional), datasets (\runsnposes, \posebusters, and \internalxtals), and metrics.
All metrics are success rate percentages where higher values are better.
The highest value across the methods for each dataset, pose aggregation, and inference regime is bolded.

On \runsnposes, we use non-\pearl metrics from the structures and poses released by the authors \cite{vskrinjar2025have}, who did not report any metrics for pocket-conditional cofolding.

On \posebusters, we use \alphafold{3}'s own released structures and metrics in the unconditional cofolding mode; we run inference and calculate metrics ourselves for \pearl and other methods.
We omit \boltz{2} from \posebusters because its structures fall within \boltz{2}'s training window. 

For \internalxtals, we run inference and calculate metrics for \pearl and other methods.

We separate \boltz{2} from the other methods and denote it with a $\dag$ due to its later \fmtdate{2023-06-01} training cutoff date, which gives it an advantage over the other baselines trained only on structures released up until \fmtdate{2021-09-30}.
\posebusters results for \af3 shown in each of the tables are based on the officially released metrics for the top pose selected by the confidence score (\ie, not computed using \bestk aggregation).

\subsection{Mixed Precision}
\label{app:mixed-precision}

Training at full precision (\texttt{fp32}) can unnecessarily limit model size due to memory and speed constraints.
A common strategy in large-scale language and protein structure models alike is to leverage half-precision arithmetic, most frequently using mixed-precision training with bfloat16 (\texttt{bf16}).
Several cofolding models such as ProteniX~\cite{bytedance2025protenix} and OpenFold~\cite{ahdritz2024openfold} adopt this approach.
Protenix adopts half-precision training aggressively, training nearly the entire model at \texttt{bf16} (with a custom CUDA \layernorm) but reverting to \texttt{fp32} for inference.
\boltz2 instead applies \texttt{bf16} only to the trunk, keeping the atom attention encoder and diffusion module in \texttt{fp32}. Our approach is to improve scalability through a conservative mixed-precision implementation designed to balance computational efficiency with numerical stability.

For training, \pearl executes most of the computationally heavy trunk module in bfloat16 (\texttt{bf16}) precision. However, numerically sensitive components, such as the loss calculations and final coordinate projections, are kept in \texttt{fp32} to ensure training stability. This strategy yields significant efficiency gains while avoiding the numerical instability issues faced by more aggressive half-precision approaches.

Specifically, we automatically upcast numerically unstable operations (\eg\ \texttt{softmax}, \texttt{exp}, \texttt{log}) to \texttt{fp32}, while computationally heavy operations (\eg\ matrix multiplications, linear projections, element-wise activations) are executed in \texttt{bf16}. All model weights are stored in \texttt{fp32}, ensuring compatibility with both \texttt{bf16} and \texttt{fp32} inference. \pearl combines these approaches: most of the trunk (triangle operations, via NVIDIA \cuequivariance kernels, and \layernorm via a custom CUDA kernel) is executed in \texttt{bf16}, while numerically sensitive modules (\eg, conditioned transitions, input and output coordinate projections, losses) are kept in \texttt{fp32}. This conservative strategy avoids instable activations, while still yielding significant efficiency gains.

\clearpage
\section{Glossary}
\label{app:glossary}

\begin{longtable}{@{}p{0.25\textwidth} p{0.75\textwidth}@{}}

\toprule
\textbf{Term / Abbreviation} & \textbf{Description} \\
\midrule
\endfirsthead

\toprule
\textbf{Term / Abbreviation} & \textbf{Description} \\
\midrule
\endhead

\midrule
\multicolumn{2}{r}{\textit{Continued on next page}} \\
\bottomrule
\endfoot

\bottomrule
\endlastfoot

AF2 & AlphaFold 2~\cite{af2} \\
AF3 & AlphaFold 3~\cite{abramson2024accurate} \\
AFDB & AlphaFold Database~\cite{varadi2024alphafold} \\
\AA & Angstrom, $10^{-10}$~m \\
apo & A protein conformation in the absence of a binding cofactor \\
\texttt{bf16} & Representation standard for floating point numbers using 16 bits \\
EqT & Equivariant transformer \\
Equivariance & The property of a function $f(x)$ such that a transformation of its input results in an equivalent transformation of its output \\
\texttt{fp32} & Representation standard for floating point numbers using 32 bits \\
holo & A ligand-bound protein conformational state \\
lDDT-PLI & Local difference distance test~\cite{Mariani2013lddt} of the protein-ligand interface \\
MSA & Multiple Sequence Alignment \\
PDB & Protein Data Bank~\cite{berman2000protein} \\
RMSD & Root mean square deviation \\
SAR & Structure-activity relationship \\
SBDD & Structure-based drug design \\
SO(3) & The group of all rotations about the origin in 3D space \\
\end{longtable}

\end{document}